# Enhanced geometry prediction in laser directed energy deposition using meta-learning

Abdul Malik Al Mardhouf Al Saadi, Amrita Basak*

* Correspond to Amrita Basak. Email address: aub1526@psu.edu

Department of Mechanical Engineering, The Pennsylvania State University, University Park, Pennsylvania 16802, USA

**Abstract**

Accurate bead geometry prediction in laser-directed energy deposition (L-DED) is often hindered by the scarcity and heterogeneity of experimental datasets collected under different materials, machine configurations, and process parameters. To address this challenge, a cross-dataset knowledge transfer model based on meta-learning for predicting deposited track geometry in L-DED is proposed. Specifically, two gradient-based meta-learning algorithms, i.e., Model-Agnostic Meta-Learning (MAML) and Reptile, are investigated to enable rapid adaptation to new deposition conditions with limited data. The proposed framework is performed using multiple experimental datasets compiled from peer-reviewed literature and in-house experiments and evaluated across powder-fed, wire-fed, and hybrid wire–powder L-DED processes. Results show that both MAML and Reptile achieve accurate bead height predictions on unseen target tasks using as few as three to nine training examples, consistently outperforming conventional feedforward neural networks trained under comparable data constraints. Across multiple target tasks representing different printing conditions, the meta-learning models achieve strong generalization performance, with $R^2$ values reaching up to approximately 0.9 and mean absolute errors between 0.03–0.08 mm, demonstrating effective knowledge transfer across heterogeneous L-DED settings.

## 1. Introduction

Laser-directed energy deposition (L-DED) is an additive manufacturing (AM) technique that enables the fabrication and repair of metal components through precise delivery of focused thermal energy and feedstock material [1], [2], [3]. Owing to its flexibility and ability to process a wide range of alloys, L-DED has become increasingly attractive for producing large scale, functionally graded, and geometrically complex parts in aerospace [4], [5], energy [6], [7], and tooling industries [8], [9]. The selection of proper process parameters (e.g. laser power, scan speed, and feedstock feed rate) is crucial for achieving geometrically accurate and defect-free builds, as these parameters directly influence the heat input, melt pool dynamics, and resulting deposition quality [10], [11], [12], [13]. However, achieving consistent geometric accuracy remains challenging due to the complex, nonlinear interactions between process parameters and the underlying thermophysical behavior of the melt pool which governs track geometry [14], [15], [16]. Therefore, accurate prediction of track geometrical features, including bead height, bead width, and cross-sectional area, based on input process parameters is essential for effective process control [17], [18]. Yet, the development of reliable predictive models is hindered by data scarcity and variability in process responses across machines, materials, and experimental setups [19], [20], [21].

Traditional approaches for identifying optimal process parameters, such as experimental trial-and-error, are often impractical, as they require extensive material consumption, prolonged experimental workflow, and substantial cost [19]. Physics-based numerical models, including finite element and computational fluid dynamics simulations, can provide detailed insight into melt pool temperature distributions, fluid flow behavior, and solidification patterns [15], [22], [23]. However, these models are computationally intensive, requiring hours to days of runtime for a single parameter set, making them unsuitable for rapid parameter exploration or process optimization. Machine learning (ML) has emerged as a powerful data-driven approach for capturing complex, nonlinear relationships between L-DED process parameters and track geometry [19], [24], [25]. Unlike physics-based models, which attempt to explicitly solve the underlying physics governing the deposition and solidification process, ML methods focus solely on learning patterns from data without explicitly solving these underlying physical processes. Nevertheless, they offer a practical and computationally efficient means of predicting process outcomes without the substantial computational cost associated with high-fidelity simulations. Once trained, ML models can generate predictions instantly, enabling rapid approximation of process outcomes. Therefore, ML techniques have gained increasing attention for various AM-related tasks, including the prediction of melt pool characteristics [24], microstructure [26], and mechanical properties [27].

Supervised learning approaches have been the most prevalent in early efforts to predict bead geometry [19] ,[28]. Classical regression models, including multiple linear regression (MLR), support vector machines (SVM), and random forests, have been widely explored due to their simplicity, interpretability, and relatively low data requirements compared to deep learning methods [29], [30] [31]. For example, early studies employed MLR to estimate bead width, bead height, and penetration depth from process parameters [32], [33], [34], [35], [36], [37]. However, the linear nature of these models limited their ability to capture nonlinear melt pool behavior [29], [38], [39]. Subsequent work advanced toward kernel-based regression models such as SVR and Gaussian Process Regression (GPR), which can capture nonlinear relationships. These models demonstrated improved prediction accuracy by accommodating nonlinear parameter interactions [40], [41], [42]. For instance, Zhu et al [43] showed that SVM predictions of bead width, height, and penetration depth improve substantially when nonlinear kernels (e.g., polynomial or RBF) are used instead of a linear kernel. Similarly, GPR has shown exceptional capability in predicting melt pool geometry through integrating multi-fidelity data generated from analytical and finite element models while also providing uncertainty quantification, as demonstrated by Menon et al [44]. Despite these successes, classical regression models often fail to generalize across different machines, materials, and parameter regimes, as they rely heavily on the statistical characteristics of the training dataset.

With the growing availability of in-situ sensing and monitoring systems, deep learning methods have gained traction for capturing complex process–response relationships beyond the capability of classical regression models [45]. Convolutional Neural Networks (CNNs), in particular, have been widely adopted for interpreting thermal images and melt pool videos [46], [47], [48]. For example, Jamnikar et al. [49] developed a multi-modality CNN that integrates melt pool images with in-situ real-time temperature measurements to predict bead height, width, penetration depth, and cross-sectional area of the entire bead including the clad and fusion zone in a wire-fed L-DED system. Their comparison with an SVR model showed mixed results, with SVR achieving lower prediction error for bead width and penetration depth, while the CNN performed significantly better for cross-sectional area and exhibited lower prediction variance across all geometric parameters. Recurrent neural networks (RNNs) have been also used to leverage the sequential nature of the deposition process by learning temporal dependencies in melt pool evolution [50], [51]. Wu et al. [52] developed an RNN-based surrogate model using LSTM, Bi-LSTM, and GRU architectures to predict melt pool peak temperature as well as melt pool length, width, and depth from time series data generated via experimentally validated multi-physics simulations. Their results showed that RNNs, particularly Bi-LSTM and GRU networks, achieve high predictive accuracy, with $R^2$ values up to 0.98 for peak temperature and above 0.88 for geometric dimensions. Despite their promising capabilities, deep learning models remain data-hungry and require substantial, domain-specific training datasets, which are costly to obtain in L-DED. As a result, the generalization challenges observed in classical regression methods persist.

To alleviate these limitations, recent studies have explored transfer learning and domain adaptation strategies to improve data efficiency and enable predictive models to generalize across different L-DED conditions. In this approach, knowledge learned from a data-rich source domain is reused to accelerate learning in a data-scarce target domain, reducing the need for extensive retraining. For example, Menon et al. [53] demonstrated that process mapping knowledge from SS316L can be effectively transferred to IN718 by fine-tuning a subset of layers in a pretrained neural network, using only 10% of the data required for training from scratch. Building upon this idea, Huang et al. [54] developed probabilistic transfer learning frameworks based on GRP to transfer melt pool geometry knowledge across three alloys in L-DED. Their models improved melt pool prediction accuracy by 22–40% when transferring knowledge from SS316L to IN718 and IN625. More recently, Huang et al. [55] extended these concepts across feedstock types, showing that transfer learning models can transfer melt pool geometry knowledge between wire- and powder-based L-DED systems. These studies collectively show that transfer learning can significantly reduce data requirements and accelerate model development for new alloys and deposition modes. However, the effectiveness of transfer learning remains highly dependent on the similarity between the source and target domains. When domain shifts are large, as is common in L-DED due to differences in thermal behavior, laser–material interaction, machine setup, and parameter ranges, transferred models exhibit limited generalization and often require substantial recalibration [21]. Moreover, collecting new data for each configuration remains costly and time consuming, restricting the scalability of data-driven models across various printing settings. These challenges highlight the need for learning frameworks that can rapidly adapt to new L-DED conditions with minimal labeled samples, motivating the adoption of few-shot learning strategies.

Meta-learning, also known as "learn to learn", is a machine learning subfield concerned with developing self-adapting models which solve a new task by leveraging experiences of solving similar related tasks [56]. Rather than training a model to perform well on a single dataset, meta-learning seeks to optimize adaptability, producing model parameters that can be fine-tuned to a new task using only a few labeled examples. This capability has made meta-learning an increasingly attractive solution for data-scarce manufacturing problems, where operating conditions, materials, and system configurations vary widely. Recent studies in advanced manufacturing have begun exploring meta-learning as a method of improving generalization of a model across various domains. For instance, Mo et al. [57] developed a hybrid model-agnostic domain generalization (H-MADG) framework for tool wear prediction across variable machining conditions. Their results showed an average root mean square error (RMSE) reduction of 36.8% when tested on NASA milling data compared with conventional techniques including supervised learning and transfer learning, demonstrating the potential of meta-learning for handling severe domain shifts in industrial settings.

Among gradient-based meta-learning methods, Model-Agnostic Meta-Learning (MAML) has gained particular prominence. Introduced by Finn et al. [58], MAML learns an initialization that can be adapted to new tasks using a single or few gradient steps and is compatible with any model trained by gradient descent, including ANN, CNN, and RNN architectures. These properties make MAML especially suitable for bead geometry prediction in L-DED, where both nonlinear spatial features (e.g., thermal images) and temporal process histories may be informative. However, MAML can suffer from high computational cost and training instability due to its reliance on second order gradient updates. To address these limitations, first order variants have been proposed, among which Reptile has gained attention for its simplicity and robustness [59]. Reptile eliminates the need for second order derivatives, resulting in improved training stability and lower computational overhead while retaining strong few-shot adaptation performance.

In metal AM, Chen et al. [60] implemented MAML to evaluate its ability to generalize across multiple synthesized bead geometry prediction tasks in L-DED. Their results showed that MAML can improve predictive accuracy when only a few labeled samples are available for a new task, outperforming conventional regression models under few-shot conditions. However, the tasks used for meta-training were synthesized by perturbing a single experimental dataset using additive and multiplicative noise modeled with GPR. Although this approach provides a practical means of constructing multiple tasks from limited data, the induced variability does not fully reflect the diversity encountered in real L-DED applications, such as differences in feedstock materials, optical configurations, depositing conditions, or process parameter regimes. As a result, the learned meta-model is not exposed to the breadth of domain shifts typical of L-DED, limiting its ability to generalize to truly heterogeneous operating conditions. This gap motivates the need for meta-learning frameworks trained on task distributions that more faithfully capture real variations in L-DED processes, enabling robust few-shot adaptation for bead geometry prediction.

To address this challenge, this work proposes a cross-dataset knowledge transfer framework based on meta-learning for geometry prediction in L-DED. The key idea is to train a meta-learner that captures transferable process–geometry relationships from multiple source datasets and rapidly adapts to a new target dataset with limited samples. By leveraging gradient-based meta-learning, the proposed approach aims to bridge the domain gap between different L-DED conditions, improving prediction accuracy, and reducing data requirements for target tasks. The effectiveness of the framework is demonstrated across multiple experimental datasets that span variations in materials, feedstock types, and processing conditions, highlighting its potential to overcome the limitations of conventional supervised learning techniques in data-scarce L-DED environments.

The main contributions of this study are as follows:
- Development of a meta-learning–based knowledge transfer framework for bead geometry prediction in L-DED across heterogeneous datasets, including both powder and wire as feedstock materials.
- Systematic analysis of transferability and adaptation performance under varying levels of available data.
- Comparative evaluation against conventional machine learning and domain adaptation methods, highlighting the benefits of meta-learning for low-data L-DED applications.
- First systematic comparison of MAML and Reptile for L-DED geometry prediction across three deposition methods.

The remainder of this paper is organized as follows. Section 2 describes the experimental methodology and data preparation; Section 3 presents the development of the meta-learning framework; Section 4 reports and discusses the experimental results; and Section 5 concludes the paper with key findings and future research directions.

## 2. Experimental method

In this paper, experimental L-DED data are gathered from multiple peer-reviewed literature sources and in-house experiments to construct a diverse and representative dataset for geometry prediction. The in-house data includes single-layer single-track depositions involving IN718 wire, SS316L wire, and IN718 wire with SS316L powder. The literature datasets include experiments conducted using various laser systems, process parameters, and alloys (i.e., SS316L, IN718, and IN738LC) enabling cross-domain evaluation of the proposed meta-learning framework. Only publications reporting complete sets of process parameters and corresponding geometric measurements are considered to ensure consistency and reproducibility of the modeling process.

The input process parameters include laser power, scan speed, powder feed rate, and wire feed rate. These variables predominantly influence the deposited track geometry and are therefore abundantly available. Other processing conditions such as shielding gas flow rate, stand-off distance, and substrate material are kept constant within each dataset but differed across datasets, reflecting variations in experimental setups among the literature sources. The resulting geometric output used in this study is the bead height, which is obtained directly from reported experimental results or digitized from graphical data when necessary. All input parameters and output measurements are converted to consistent units prior to model development. To enable meta-learning across diverse experimental conditions, the collected data are partitioned into multiple tasks, each corresponding to a dataset obtained from a single source representing specific combinations of material type, laser system, or deposition strategy. Organizing the data in this manner allows the meta-model to extract shared knowledge across distinct yet related tasks and evaluate its adaptability to new, unseen tasks. For instance, IN718 deposition data from one study may serve as the target domain, while the remaining datasets, including those for SS316L and IN718, serve collectively as the source domain for meta-training.

Prior to model training, data preprocessing steps are applied uniformly across all datasets. Input variables are normalized using fixed minimum and maximum bounds defined from domain knowledge, ensuring consistent scaling across tasks while avoiding information leakage from statistics of individual tasks. This approach is particularly important in both meta-training and meta-testing, where only a small subset of data from each task is provided to the model to learn task-specific knowledge. Normalizing based on such limited samples could lead to inconsistent scaling, whereas normalizing using the entire task exposes information from the data reserved for evaluation. The datasets are then split into training, validation, and testing partitions for consistent model evaluation. The training and validation data from multiple sources are used to train the meta-model, while the target domain test data are reserved exclusively for evaluating adaptation performance under limited data conditions. Random search is employed to optimize model hyperparameters, with k-fold cross-validation performed during tuning to prevent overfitting the optimized model to either the validation or the test set. This literature-based data compilation approach enables the development and assessment of a generalizable predictive model without the need for new experimental campaigns, while still capturing the wide variability inherent in L-DED processes.

## 3. Meta-model development

The development of the proposed meta-model involves two key stages: (1) organizing the available data into task formulations suitable for a meta-learning training framework and (2) designing a neural network-based architecture capable of adapting to different datasets through meta-learning algorithms. In this study, the base model is a simple feedforward (vanilla) neural network, which serves as the fundamental model for both MAML and Reptile implementations. The goal is to enable fast adaptation to new L-DED conditions with minimal retraining effort.

### 3.1 Task generation

To enable meta-learning, the compiled L-DED datasets are divided into a set of tasks, each representing a distinct process or experimental condition. A task is defined as a mapping from process parameters such as laser power, scanning speed, powder feed rate, and wire feed rate to bead height. Each dataset from the literature corresponding to a specific material, system, or deposition setup, is treated as an independent task. In this study, 14 tasks are used, comprising seven LP-DED (powder-only) tasks, six LW-DED (wire-only) tasks, and one LWP-DED (wire-powder) task. Table 1 summarizes all tasks, while Fig. 1 illustrates the distribution of bead height values for each group of tasks. From Fig. 1, it is noticeable that powder-only tasks extend over a wider height range covering the region from 0.1 mm to 2.2 mm. Wire-only bead height values distribution, excluding Task 4, is contained within a smaller range between 0.4 mm and 1.2 mm. These variations in tasks distributions can pose a challenge to the transferability knowledge across tasks which will be further discussed in section 4.

**Table 1:** Summary of L-DED tasks used in this study, including literature sources for each task.

| Task number | Feedstock form | Feedstock material | Number of samples | Reference |
|---|---|---|---|---|
| Task 1 | Powder | SS316L | 25 | [61] |
| Task 2 | Powder | IN718 | 27 | [62] |
| Task 3 | Powder | IN738LC | 13 | [63] |
| Task 4 | Wire | SS316L | 22 | [64] |

| Task 5 | Wire | SS316L | 44 | In-house data |
| Task 6 | Wire | IN718 | 16 | [65] (Feed angle 2.5 degrees) |
| Task 7 | Wire-powder | IN718 wire and SS316L powder | 36 | In-house data |
| Task 8 | Wire | IN718 | 6 | [65] (Feed angle 0 degrees) |
| Task 9 | Wire | IN718 | 5 | [65] (Feed angle 7.5 degrees) |
| Task 10 | Powder | IN718 | 24 | [66] |
| Task 11 | Powder | IN718 | 9 | [67] |
| Task 12 | Powder | SS316L | 9 | [68] |
| Task 13 | Powder | IN718 | 25 | [69] |
| Task 14 | Wire | IN718 | 9 | In-house data |

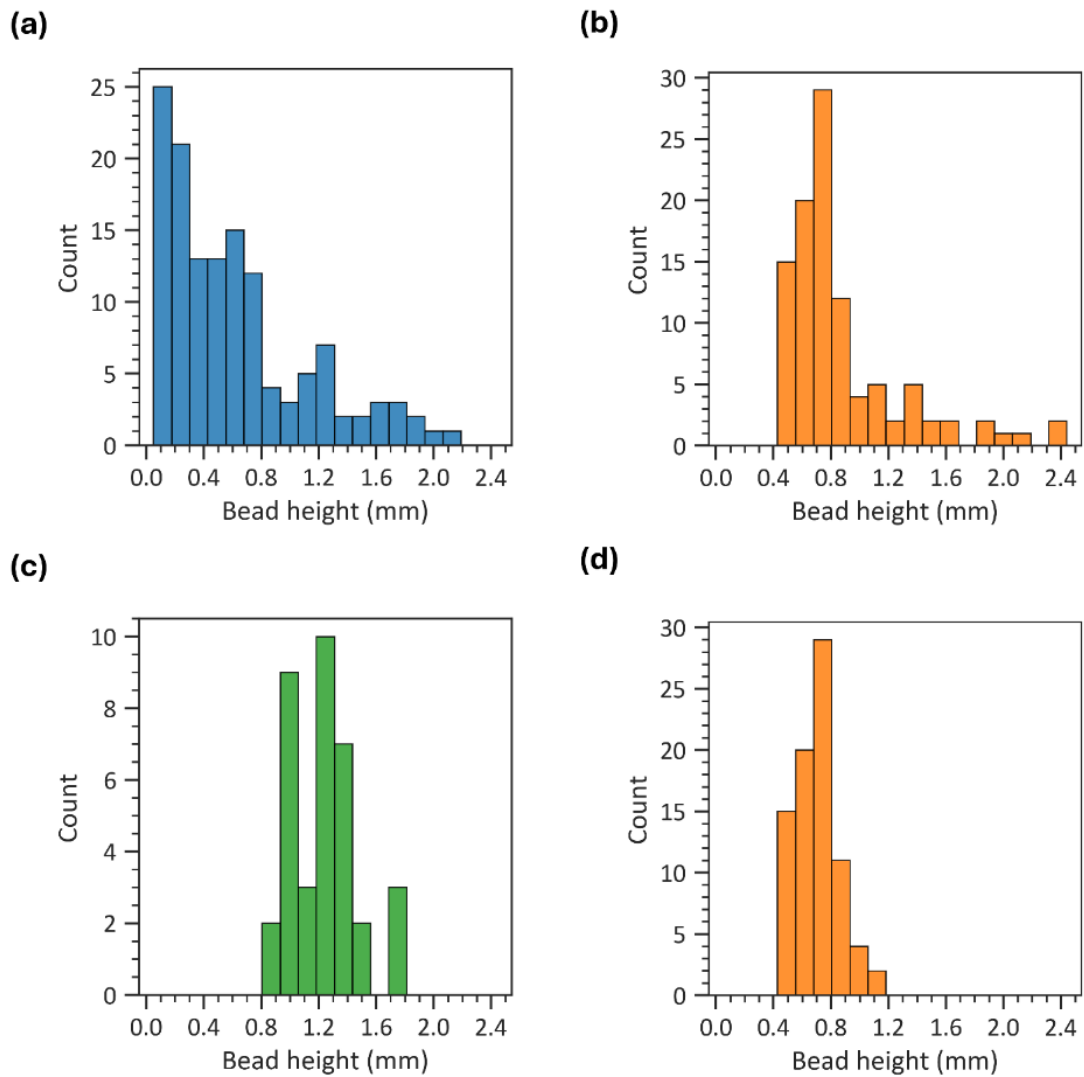

**Fig. 1.** Distribution of bead height values for (a) powder-only L-DED tasks, (b) wire-only L-DED tasks, (c) wire-powder L-DED tasks, (d) wire-only L-DED tasks excluding Task 4.

At the beginning of the meta-learning process, a task generator is implemented to create internal training and validation splits within each task. The training split, later referred to as the support set, is used for task-specific adaptation in the inner loop, while the validation split, or query set, is used in the outer loop to evaluate the performance and update the meta parameters of the generalizable model. This task sampling strategy ensures that the meta-model is exposed to a wide range of process–geometry relationships and learns an initialization that generalizes well across different domains. Since the number of samples varies across tasks, the support set for each task is defined to include a specific percentage (20% for baseline performance) of randomly selected samples, while the remaining data are assigned to the query set. To ensure a reliable evaluation of task-specific adaptation, the support set of the target task is resampled five times and the average performance is reported in this study. This method of evaluation considers the variations in the model's performance due to variations in the support set. In the meta-learning framework, the tasks are organized into three groups: training, validation, and testing. The testing group corresponds to the target task used to assess generalization performance, while the remaining tasks form the meta-training pool. During meta-training, a 5-fold validation strategy is employed, where two tasks are randomly selected as the validation group, and the rest are used for training in each fold. All random sampling procedures are performed using a fixed random seed to ensure consistent data splits and reproducible results across experiments.

### 3.2 Creation of a feedforward neural network

A feedforward neural network (NN) is designed as the base learner for all meta-learning experiments. The model architecture consists of an input layer corresponding to the process parameters, three hidden layers with 64 neurons per layer, and an output layer predicting the bead height. The network employs hyperbolic tangent (Tanh) activation functions to introduce nonlinearity and mean squared error (MSE) as the loss function. The model parameters are initialized randomly and updated using gradient-based optimization. The vanilla NN is trained using both conventional supervised learning to establish a baseline prediction accuracy for geometry estimation, and within a meta-learning framework to evaluate the benefits of meta-learning algorithms.

### 3.3 Meta learning architecture

Meta-learning architectures are designed to enable models to learn how to learn, allowing them to quickly adapt to new tasks. Unlike conventional transfer learning, meta-learning focuses on learning the adaptation mechanism itself, enabling rapid generalization to unseen tasks with limited data and minimal manual tuning. The ability of meta-learning algorithms to self-adapt is attributed to its unique training setup where the model learns in two steps: meta-training, and meta-testing.

The meta-training step is where the model learns how to learn, while the meta-testing step is where the model learns how to adapt to unseen tasks. The meta-training step is designed as a bilevel optimization problem, where the inner objective (inner loop) aims to adapt the model's parameters (weights and biases ($\theta$)) to each specific task in the training set individually using limited task-specific data, while the outer loop objective is to optimize the initial parameters ($\theta_o$) such that the adapted model performs well across all tasks in the training set.

Fig. 2 illustrates the bilevel optimization process during meta-training, where task-specific inner loop updates are followed by an outer loop meta-update to learn a shared initialization across tasks. Here, a task ($T_i$) represents a dataset sampled from a data distribution ($P(T)$). Each task is distributed into a support set ($T_{i,S}$) and a query set ($T_{i,Q}$). The $T_{i,S}$ is used to train $\theta$ to adapt to specific tasks in the inner loop, while the $T_{i,Q}$ is used to guide the optimization of $\theta_o$. In the meta-testing step, the $T_{i,S}$ is used to fine-tune $\theta_o$, which is then evaluated on the unseen $T_{i,Q}$. In this work, two representative gradient-based algorithms, MAML and Reptile, are used to develop a generalizable model capable of rapidly adapting to new deposition conditions with minimal data samples and fine-tuning requirements.

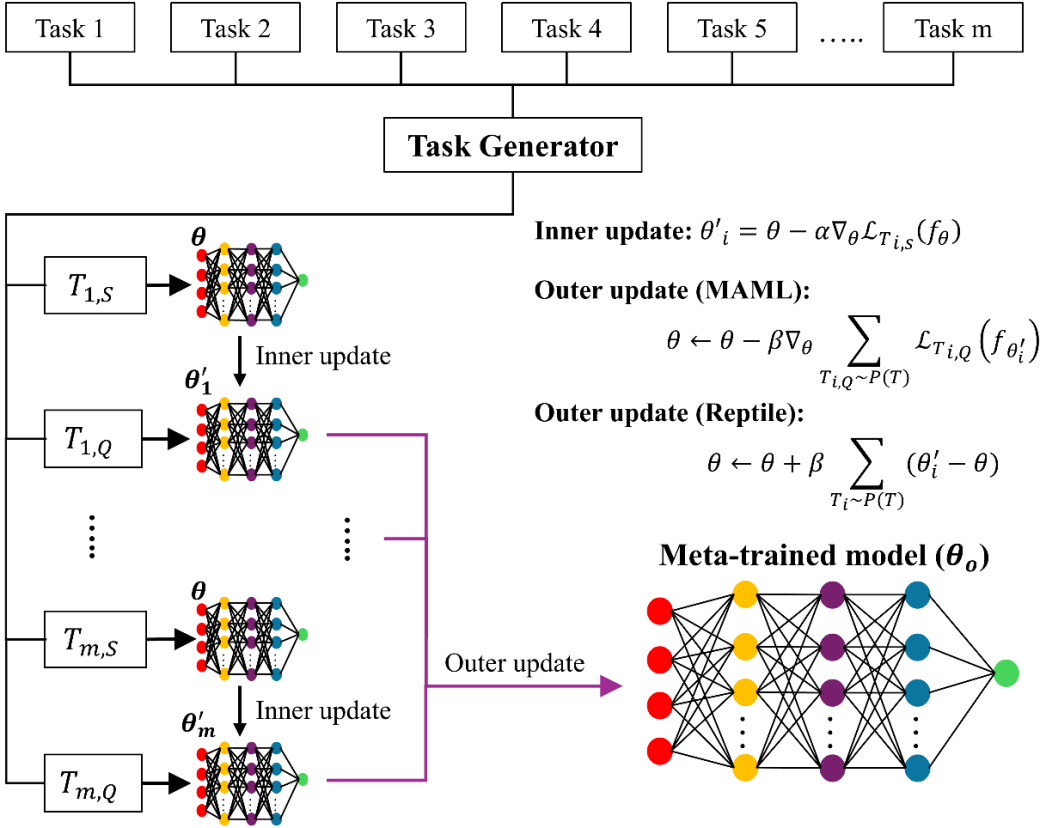

**Fig. 2.** Schematic illustration of the meta-training framework used in this study.

### 3.3.1 Model-agnostic Meta-Learning (MAML)

MAML is a gradient-based meta-learning algorithm, which utilizes stochastic gradient descent (SGD) to optimize and further adapt $\theta_o$. It is referred to as model-agnostic because it is applicable to any model that is trained using gradient descent to minimize its loss function. MAML follows the same training setup described in Section 3.3, optimizing both the inner and outer loop objectives through SGD. During the meta-training step, a model ($f_\theta$) is trained on the support sets of each training task by performing at least one gradient step per task in the inner loop as follows:

$$\theta'_i = \theta - \alpha \nabla_\theta \mathcal{L}_{T_{i,S}}(f_\theta), \tag{1}$$

where $\mathcal{L}_{T_{i,S}}(f_\theta)$ is the loss function computed on the support set $T_{i,S}$, $\alpha$ is the inner loop learning rate (step size), and $\theta'_i$ represents the adapted parameters after one gradient update on $T_{i,S}$.

Once the inner loop updates are completed for all training tasks, the model is evaluated on the corresponding query sets $T_{i,Q}$, and the meta parameters are updated in the outer loop as follows:

$$\theta \leftarrow \theta - \beta \nabla_\theta \sum_{T_{i,Q} \sim P(T)} \mathcal{L}_{T_{i,Q}}\left(f_{\theta'_i}\right), \tag{2}$$

where $\beta$ denotes the learning rate for the outer loop, $\mathcal{L}_{T_{i,Q}}\left(f_{\theta'_i}\right)$ is the loss function evaluated on the query set $T_{i,Q}$ using the adapted parameters for $T_{i,S}$ obtained from the inner loop. The objective of this optimization framework is to reach $\theta_o$ which leads to a good performance on any target task after $x$ inner loop gradient steps on that task. The gradient of the outer loop objective is computed with respect to $\theta$, requiring the application of the chain rule through the inner update:

$$\nabla_\theta \sum_{T_{i,Q} \sim P(T)} \mathcal{L}_{T_{i,Q}}\left(f_{\theta_i'}\right) = \frac{\partial \sum \mathcal{L}(f_{\theta'})}{\partial \theta'} \cdot \frac{\partial \theta'}{\partial \theta}. \tag{3}$$

This dependency makes MAML a second order optimization algorithm, requiring Hessian-vector computations. Such operations are supported by modern deep learning frameworks, including TensorFlow and PyTorch. Once the meta-training phase is completed after running a sufficient number of epochs, the optimized initialization ($f_{\theta_o}$) is expected to adapt rapidly to any unseen task sampled from $P(T)$ using only a few gradient updates on its corresponding $T_{i,S}$.

In this study, MAML is trained by repeatedly sampling a batch of L-DED tasks at each meta-iteration, performing one inner loop update on the support set of each task, and applying a single outer loop update based on the corresponding query losses. The resulting meta parameters serve as the initialization from which the model is fine-tuned to target L-DED tasks during meta-testing using only a few support samples. This initialization is expected to enable rapid adaptation across a range of unseen task types, including powder-only, wire-only, and wire–powder L-DED conditions.

### 3.3.2 Meta-learning with Reptile

Reptile is a first order gradient-based meta-learning algorithm that shares conceptual similarities with MAML but eliminates the need for second order gradient computations. Similar to MAML, it seeks to learn an optimal initialization $\theta_o$ that enables the model to quickly adapt to new tasks with minimal fine-tuning. However, instead of explicitly computing the meta-gradient through backpropagation across the inner and outer loops, Reptile performs multiple SGD updates within each task and moves the initialization parameters $\theta$ toward the updated model's parameters $\theta_i'$. This makes Reptile computationally simpler and more memory efficient than MAML, while retaining comparable adaptation capability. During the meta-training step, for each task $T_i$ sampled from the task distribution $P(T)$, the model parameters are updated on the support set $T_{i,S}$ for $k$ gradient steps as follows:

$$\theta_i' = \theta - \alpha \sum_{j=1}^{k} \nabla_\theta \mathcal{L}_{T_{i,S}}(f_\theta), \tag{4}$$

where $\alpha$ is the learning rate for the inner loop updates, $L_{T_{i,S}}(f_\theta)$ is the loss function computed on the support set $T_{i,S}$, and $\theta_i'$ represents the adapted parameters after $k$ updates per task. Once the inner loop adaptation is completed for all sampled tasks, the meta-update is applied to move the initialization $\theta$ closer to $\theta_i'$ according to:

$$\theta \leftarrow \theta + \beta \sum_{T_i \sim P(T)} (\theta_i' - \theta), \tag{5}$$

where $\beta$ is the learning rate for the outer update. Unlike MAML, which explicitly backpropagates through the inner update using the chain rule and thus requires Hessian-vector products, Reptile estimates the meta-gradient implicitly by measuring how the parameters change after task specific adaptation. This makes Reptile a first order approximation to MAML, significantly reducing computational cost and simplifying implementation while maintaining strong performance in few-shot and regression settings. Once the meta-training phase is completed over multiple epochs, the optimized initialization $f_{\theta_o}$ can rapidly adapt to unseen tasks drawn from $P(T)$ using only a few additional gradient steps on their respective support sets $T_{i,S}$. In this study, Reptile is trained using the same task generator and adaptation protocol as MAML to ensure direct comparability between the two algorithms.

### 3.3.3 Model training and testing
Both the MAML and Reptile models are trained following the same meta-learning workflow to ensure consistent comparison between the two algorithms. During meta-training, tasks are repeatedly sampled from the meta-training pool and divided into support and query sets according to the task generation procedures described earlier. Each task is randomly split, such that 20% of the samples forms the support set used for the inner loop adaptation, while the remaining 80% forms the query set which is used to evaluate the adapted model and compute the outer loop update. Fig. 3 illustrates the meta-testing procedure followed in this work.

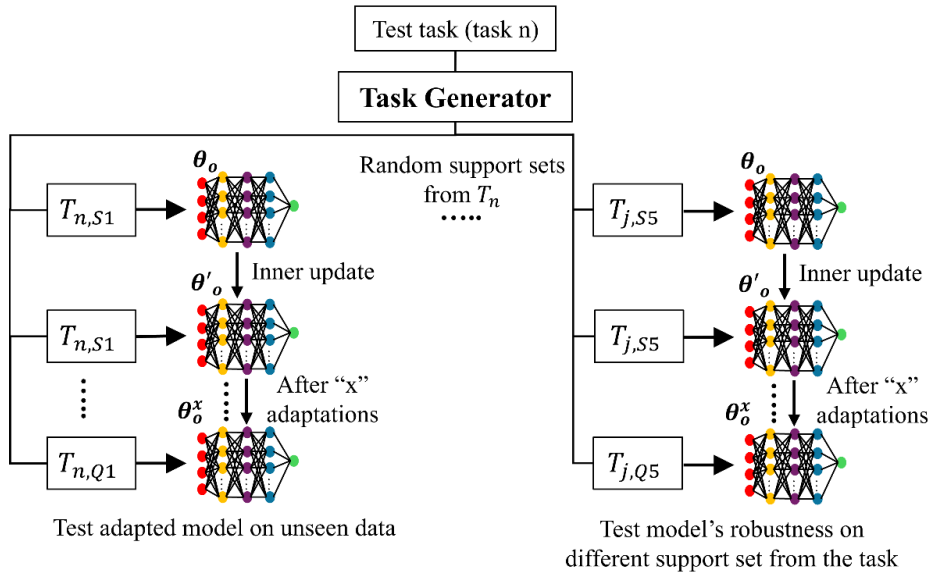

**Fig. 3.** Schematic illustration of the meta-testing procedure followed in this study.

In all meta-learning experiments, the input process parameters for every task are normalized using fixed, predefined bounds derived from domain knowledge to ensure consistent scaling across diverse datasets and to avoid information leakage [70]. Specifically, laser power, scan speed, powder feed rate, and wire feed rate are each normalized using global maximum values of 3000 W, 2000 mm/min, 25 g/min, and 10 g/min, respectively. Using fixed bounds rather than normalizing tasks separately prevents information leakage from target tasks and maintains a unified parameter space across all tasks. In addition, bead height values are transformed using a $\log(1 + Height)$ mapping to reduce skewness of the data distribution and help prevent large variations from dominating the training process [71]. These preprocessing steps contribute to stable model training and ensure consistent model behavior across the diverse L-DED tasks used in this study.

Hyperparameters governing the meta-learning process, including the inner loop learning rate, outer loop learning rate, and batch size, are selected using Optuna with a random search strategy [72]. To ensure generalizability and prevent the hyperparameters from overfitting to any specific dataset, a 5-fold cross-validation approach is employed, where two tasks are held out as validation tasks in each fold and the remaining tasks are used for meta-training. This procedure yielded an optimal inner loop learning rate of 0.1 and an outer loop learning rate of 0.002. All models are trained for 100 epochs. When training on the full set of tasks, a batch size of 6 is used for all models. However, when training tasks are restricted to wire-only or powder-only subsets, the batch size is manually adjusted to account for the reduced number of available tasks. Aside from these batch size adjustments, all remaining hyperparameters are fixed across experiments. The number of inner and outer loop adaptation steps used during evaluation are reported for each experiment in Section 4.

## 4. Results and discussion
### 4.1 Validation with literature results
Before applying the proposed meta-learning framework to the L-DED datasets compiled in this study, it is necessary to verify the correctness and stability of the implemented algorithms. The MAML and Reptile codes used herein are adapted from a publicly reproduced TensorFlow 2.0 implementation of Finn et al.'s sinusoidal regression example [73].

The algorithms are first validated by reproducing the benchmark sinusoidal meta-learning task presented by Finn et al., ensuring that the core inner and outer optimization structures are correctly implemented for the meta-training step as well as the task-specific adaptation and evaluation structures for the meta-testing step. Subsequently, the MAML algorithm is used to reproduce the bead geometry prediction results reported in the referenced study [60] mentioned earlier in the introduction section, which employed a GPR to synthesize the training and testing tasks from a small experimental dataset. To enable this reproduction, a GPR module is developed and integrated into the meta-learning framework, following the task generation procedure described in that study. To diversify the tasks, additive and multiplicative noises are introduced into the GPR outputs, as described in the referenced study. Finally, the MAML model is constructed using the same hyperparameters reported in the study, including the neural network architecture, activation functions, learning rates, and the number of inner and outer loop updates.

For the reproduction of the sinusoidal meta-learning benchmark, each task is defined as fitting a sine function with a randomly sampled amplitude within [0.1, 5.0] and a phase sampled within $[0, \pi]$, while the model is required to adapt to a new function using only a few support samples. Using the adapted TensorFlow 2.0 implementation, the MAML and Reptile algorithms are trained using the same task distribution and neural network architecture described in the original studies, ensuring methodological consistency. The inputs (x-values) are sampled from [-5, 5] for both models. The regression model consisted of a feedforward neural network consisting of two hidden layers, each containing 40 neurons with rectified linear unit (ReLU) as the activation function. For the MAML implementation, the inner loop update is performed with a step size of 0.01, and the meta parameters are optimized using Adam optimizer. In the Reptile implementation, task adaptation is performed using an inner loop learning rate of 0.01, an outer loop learning rate of 0.001, and 32 inner updates. These settings copy the structure and learning dynamics of the original benchmark, allowing a direct assessment of whether the implemented framework reproduces the expected meta-learning behavior. Fig. 4 illustrates how MAML and Reptile perform on a test sinusoid task after different numbers of adaptation steps.

The reproduced results show close agreement with both Finn et al.'s original findings and the publicly available reproduction code. As illustrated in Fig. 4, both MAML and Reptile can fit the sinusoidal function after only a single adaptation step, with an even closer match obtained after ten steps. Notably, the algorithms demonstrate stable adaptation regardless of whether the sampled input domain spans only one side of the sine wave or extends over the entire wave, indicating robustness to variations in the support set domain. Quantitatively, the MSE achieved by MAML and Reptile in this study are 0.4 and 0.3, respectively, which are comparable to the approximate MSE of 0.4 observed from the learning curve in the original MAML study. These findings confirm that the implementations faithfully replicate the learning scheme and performance documented in prior work, thereby providing confidence in the correctness and reliability of the meta-learning framework used in this study.

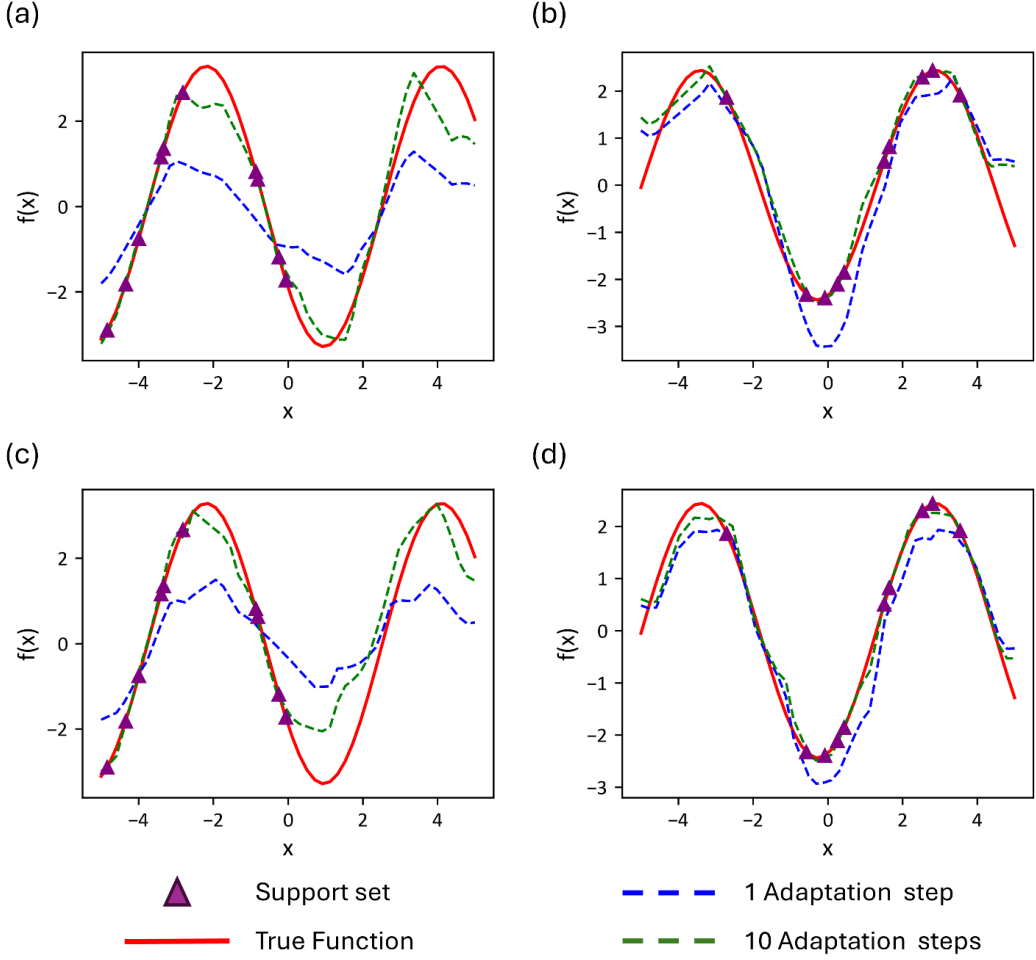

**Fig. 4.** Demonstration of MAML and Reptile performance on a sinusoidal test task using 1 and 10 meta-updates (code adapted from [73]). (a) MAML with a support set limited to one side of the sine wave. (b) MAML with a support set spanning both sides of the wave. (c) Reptile with the support set limited to one side. (d) Reptile with the support set spanning both sides.

Beyond the sinusoidal benchmark, the framework is additionally validated by reproducing the Self-Generating Multi-Task Model-Agnostic Meta-Learning (SGM-MAML) framework reported in [60]. The SGM-MAML reproduction is carried out using the same meta-learning architecture described in the original study, including a seven-layer feedforward neural network with ReLU activation, a single inner loop adaptation step with a learning rate of 0.016, and an Adam optimizer with cosine-annealing scheduling for meta parameters optimization. A total of 1000 tasks are used for meta-training, and 25 additional tasks are reserved for meta-testing. Fig. 5 presents the predicted versus true bead geometry features for a representative test task after ten adaptation steps. The reproduced SGM-MAML model achieves strong agreement between predictions and true values, as reflected by the mean Pearson correlation coefficients of 0.983 for width, 0.996 for height, and 0.997 for depth across the test tasks. These results are in close alignment with the performance reported in the original study, which documented correlation coefficients of approximately 0.985 for width, 0.996 for height, and 0.998 for depth, confirming that the implemented task generator and meta-learning algorithm reproduce the expected behavior of the reference SGM-MAML framework.

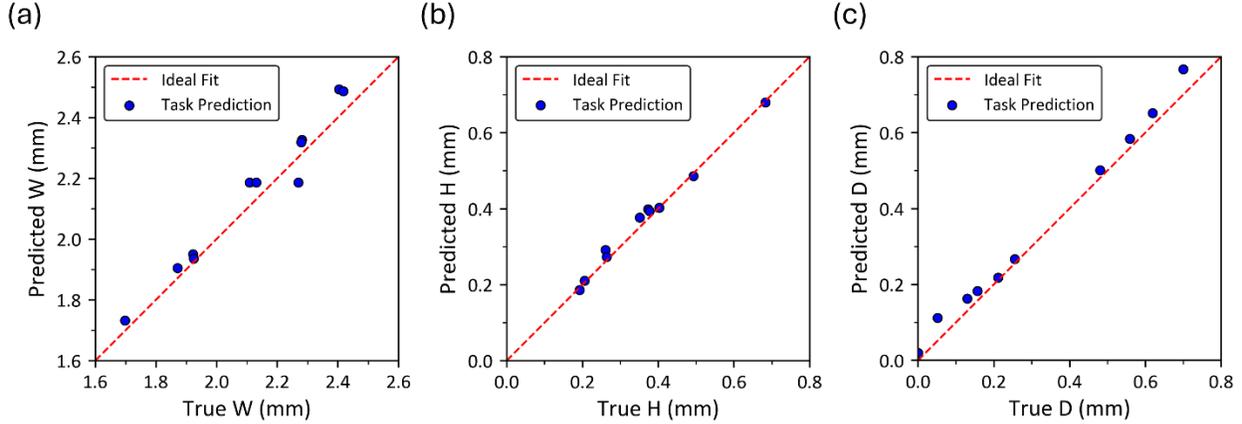

**Fig. 5.** Predicted versus true bead geometric features obtained from the reproduction of the SGM-MAML framework: (a) bead width, (b) bead height, and (c) penetration depth. Here, "W", "H", and "D" represent bead width, height, and depth, respectively.

Overall, Figs. 4 and 5 demonstrate that the implemented meta-learning framework accurately reproduces the benchmark sinusoidal and SGM-MAML results reported in the literature. The close agreement in both qualitative behavior and quantitative performance confirms that the algorithms, task generator, and testing procedures have been correctly implemented and thoroughly validated. With this verification complete, the framework can now be applied to the L-DED datasets compiled in this study.

### 4.2 Performance of MAML and Reptile for powder-only L-DED
To assess the capability of the meta-learning framework to generalize and adapt to unseen powder-based L-DED conditions, the performance of the MAML and Reptile models is evaluated using five independent shuffles of the same target task. Each shuffle corresponds to a distinct support set sampling, where 20% of the data from the test task is randomly selected for inner loop adaptation, and the remaining samples are used for evaluation. This resampling strategy mitigates variability arising from the limited task size and provides a more robust estimate of adaptation performance. Task 1, previously introduced in Table 1, is selected as the baseline target task, while tasks 2 and 3 will later be used to assess the broader generalizability of the models.

### 4.2.1 MAML performance, convergence behavior, and generalizability trends
The MAML model is first evaluated using one inner loop gradient step during meta-training and five adaptation steps during meta-testing. The MAML model achieves an average coefficient of determination ($R^2$) score of 0.84 across the five support set shuffles. Given that the target task contains only 25 samples, this result demonstrates MAML's ability to learn and adapt to an unseen process condition using only a 5-shot support set and a limited number of fine-tuning steps. Next, the MAML model is evaluated under varying numbers of inner loop gradient steps during meta-training and adaptation steps during meta-testing to assess their influence on prediction performance. Fig. 6 illustrates the predicted versus true bead height values for MAML under different combinations of inner loop gradient steps and target task adaptation steps examined in this study, including configurations with five inner loop steps paired with five, twenty, and fifty adaptation steps. Increasing the number of inner loop gradient steps during meta-training from one to five yields an improvement in average $R^2$ to 0.87. This improvement is expected as additional inner loop updates allow the meta-learner to continue updating its parameters towards better initialization parameters instead of optimizing for a single step update. As a result, the optimized parameters lie closer to a region in the parameter space that is broadly suitable for rapid adaptation across diverse L-DED tasks. The trade-off is a modest increase in computational cost, which may become significant for larger datasets, though the impact is minimal for the compact datasets used in this study. Further improvements are obtained by increasing the number of adaptation steps during meta-testing from five to twenty, resulting in an average R² score of 0.89. This improvement arises from the model continuing to adjust its parameters during meta-testing, progressively aligning with the characteristics of the target task. These additional updates enable the model to better capture the nonlinear behavior of L-DED parameters in the target task, leading to more accurate predictions.

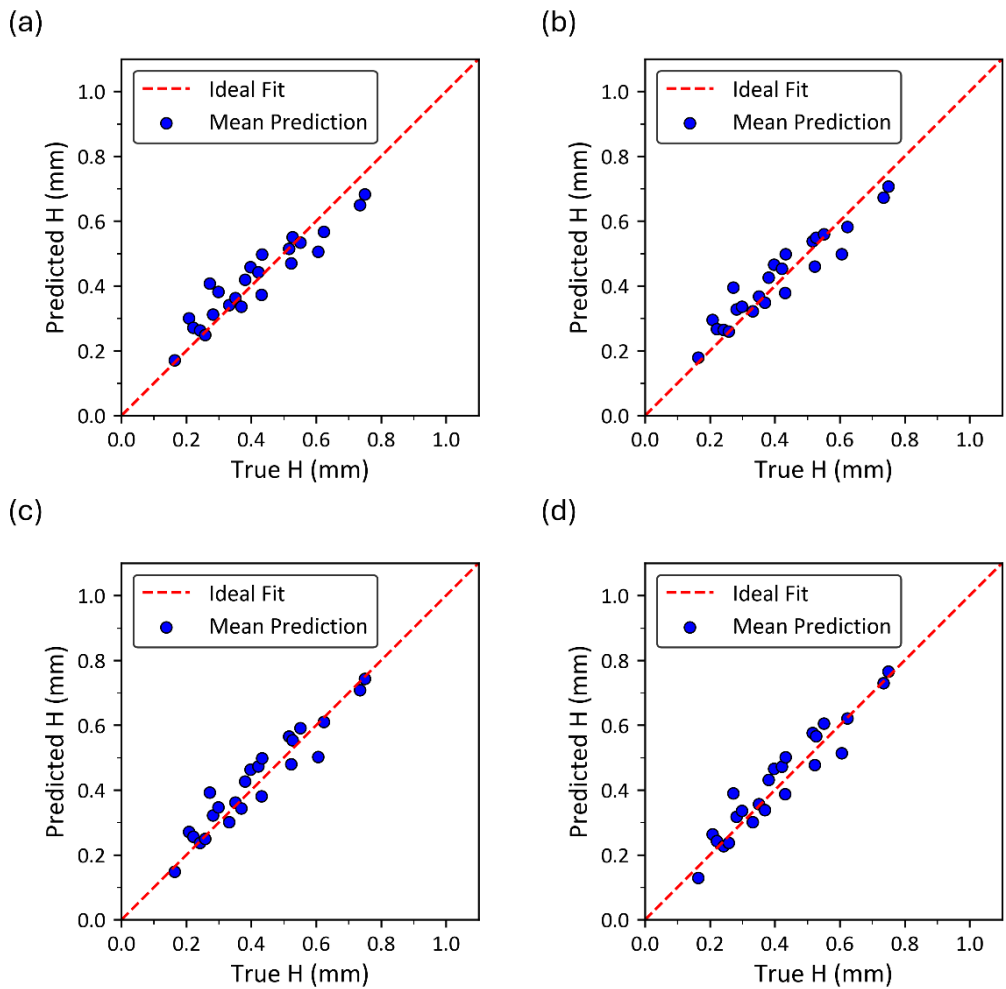

**Fig. 6.** Performance of MAML under different meta-training and meta-testing configurations. Predicted versus true bead height with (a) single inner loop step and five adaptation steps, (b) five inner loop steps and five adaptation steps, (c) five inner loop steps and twenty adaptation steps and (d) five inner loop steps and fifty adaptation steps.

The model's bead height predictions are recorded after each adaptation step to quantify its convergence behavior. These predictions are obtained without updating the model's parameters, ensuring that no information from the query set influences the adaptation process. The MSE of the model's predictions versus true bead height values is then computed at each adaptation step to assess the convergence behavior of the model toward the target task. Fig. 7 shows the average MSE across the five support set shuffles over 50 adaptation steps. Both configurations show a similar reduction in the average MSE as adaptation progresses, indicating that the model improves its predictions with additional updates. The MAML model trained with a single inner loop gradient step converges quickly, reaching stable performance after the first adaptation step, whereas the model trained with five inner loop steps continues to improve over a larger number of adaptation steps. At 50 adaptation steps, the single step configuration reaches an average MSE of 0.003 mm² (mean absolute error (MAE) of 0.045 mm), while the five steps configuration approaches an average MSE of 0.003 mm² (MAE of 0.043 mm). The shaded regions represent one standard deviation from the average values, reflecting variability across the shuffled support sets.

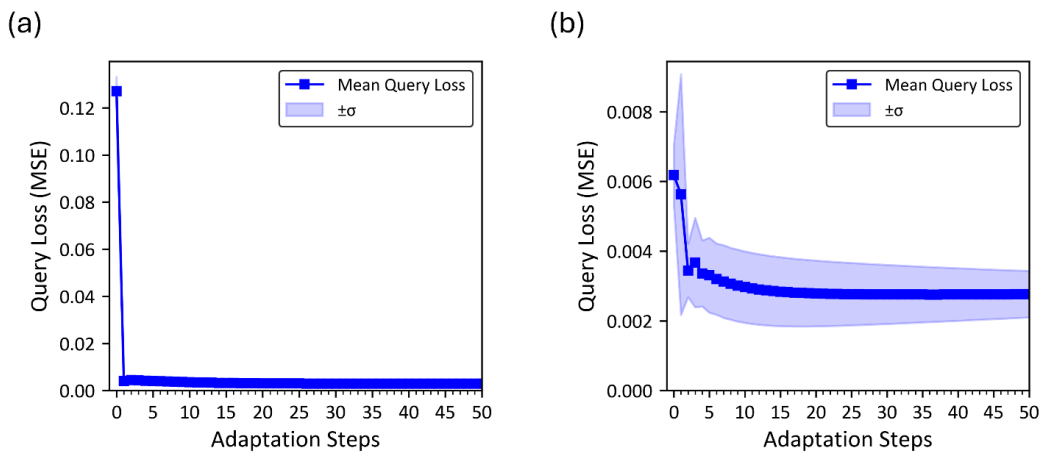

**Fig. 7.** MSE versus adaptation steps for (a) single gradient step and (b) five gradient steps.

Additionally, table 2 presents the model's performance metrics values, including Pearson's correlation factor (r), $R^2$, MSE, and MAE, for each of the five target task shuffles. Although the overall performance remains consistent across shuffles, the variation in $R^2$ and MAE indicates that the support set composition can influence adaptation accuracy. This behavior is expected in few-shot learning, where only a small number of samples guide the parameter updates during meta-testing. Consequently, careful selection of a representative support set is recommended to ensure effective adaptation and maximize predictive accuracy for new L-DED tasks.

**Table 2:** Regression performance metrics of the MAML model across five target task shuffles, including Pearson's correlation coefficient (r), coefficient of determination ($R^2$), mean squared error (MSE), and mean absolute error (MAE).

| Model Number | Pearson's correlation factor (r) | R² score | MSE (mm²) | MAE (mm) |
|---|---|---|---|---|
| Model 1 | 0.98 | 0.84 | 0.004 | 0.054 |
| Model 2 | 0.98 | 0.93 | 0.002 | 0.037 |
| Model 3 | 0.96 | 0.85 | 0.004 | 0.050 |
| Model 4 | 0.96 | 0.90 | 0.003 | 0.044 |
| Model 5 | 0.96 | 0.91 | 0.002 | 0.039 |

To further assess the generalizability of MAML on powder-only L-DED tasks, the model is retrained with the original target task moved into the meta-training set and another, previously unseen task assigned as the target. Five independent support set shuffles are again used for evaluation. The new target tasks, Tasks 2 and 3, provide an opportunity to examine how well the model adapts to process conditions not encountered during the original meta-training. As shown in Fig. 8, MAML maintains comparable performance on Task 2, achieving an average $R^2$ score of 0.82 along with MSE and MAE values of 0.004 mm² and 0.052 mm, respectively. In contrast, performance on Task 3 shows a noticeable reduction in the average $R^2$ score which is 0.24 corresponding to average MSE of 0.004 mm² and average MAE of 0.05 mm. This decrease is likely influenced by the smaller size of Task 3, which contains only 13 samples, yielding a support set of only three points under the 20% sampling protocol. Such a limited support set restricts the model's capacity to adapt to the target task during meta-testing. Increasing the support set samples to 50% of the task increased the $R^2$ score to 0.81, where the average MSE is reduced to 0.001 mm$^2$, and the average MAE is reduced to 0.027 mm.

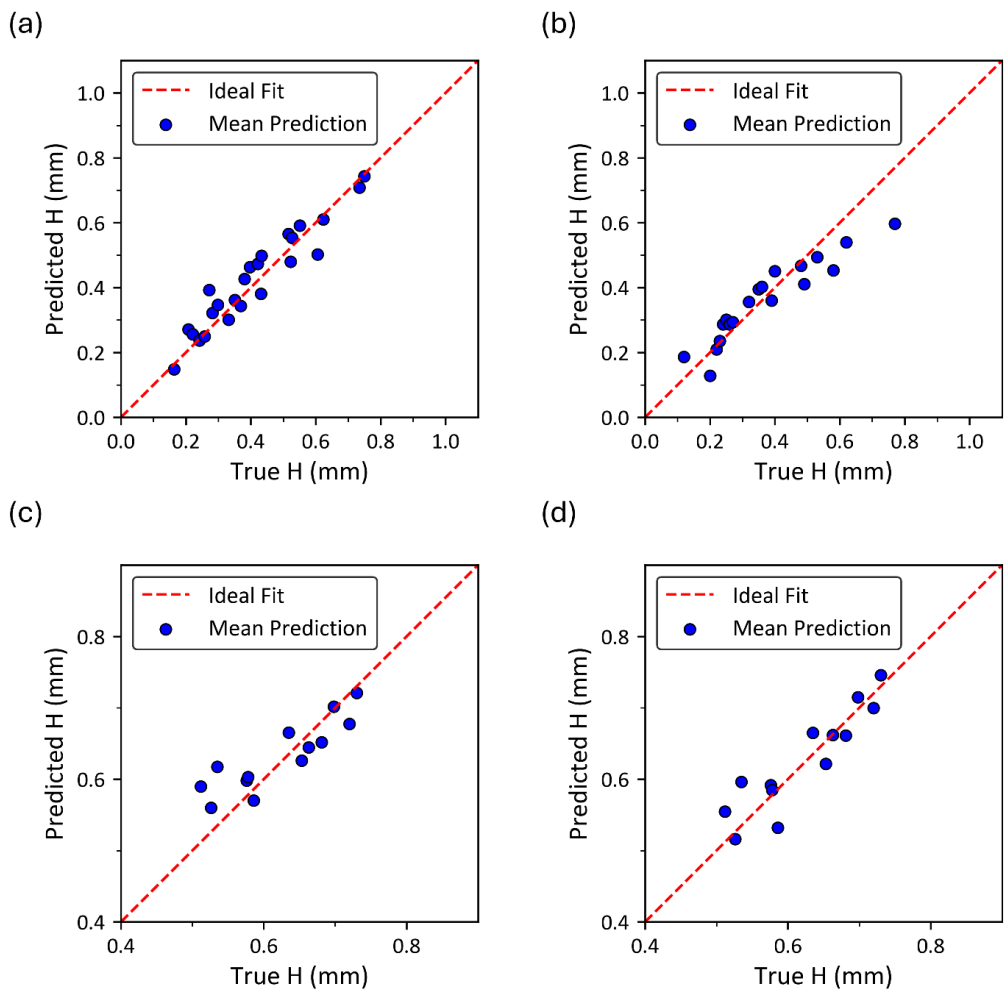

**Fig. 8.** Generalizability of MAML across different target tasks. Predicted versus true bead height for (a) Task 1 using 20% of the task data as the support set, (b) Task 2 using 20% of the task data as the support set, (c) Task 3 using 20% of the task data as the support set, and (d) Task 3 using 50% of the task data as the support set. Here, "H" represents bead height.

### 4.2.2 Reptile performance, convergence behavior, and generalizability trends

Under the baseline configuration of one inner loop gradient step and five adaptation steps, the Reptile model achieved an average $R^2$ score of 0.84, matching the baseline performance of MAML. The average predicted bead height values for the five support set shuffles versus true values under the different inner loop gradient step and adaptation step configurations are presented in Fig. 9. Increasing the number of inner loop updates during meta-training to five improved the average $R^2$ score to 0.87. In Reptile, this improvement occurs because the algorithm moves the meta-initialization toward the average of multiple tasks adaptations during the meta-training step. Additional gradient steps allow each task to explore its parameter space more thoroughly, producing a more informative direction for the outer update. As a result, the learned initialization becomes more representative of the shared characteristics across training tasks. Extending the number of adaptation steps during meta-testing from five to twenty further improved performance, yielding an average R² score of 0.89. The additional adaptation steps give the model more opportunity to specialize to the target task, further improving predictions accuracy of the target task.

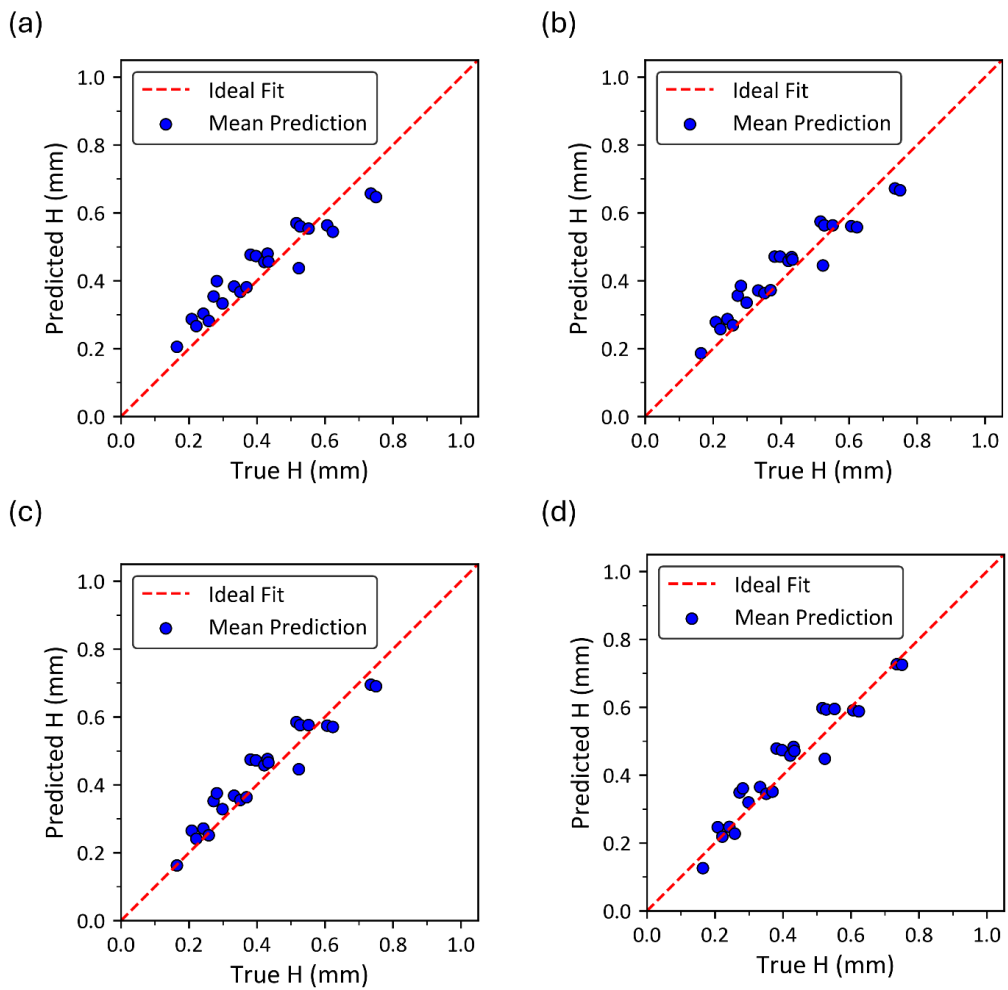

**Fig. 9.** Performance of Reptile under different meta-training and meta-testing configurations. Predicted versus true bead height with (a) single inner loop step and five adaptation steps. (b) five inner loop steps and five adaptation steps. (c) five inner loop steps and twenty adaptation steps. (d) five inner loop steps and fifty adaptation steps. Here, "H" represents bead height.

The convergence behavior of Reptile follows a similar pattern to that observed for MAML, with a sharp reduction in MSE after the first adaptation step and continued gradual improvement as additional steps are performed. Fig. 10 illustrates average MSE across the five support set shuffles over 50 adaptation steps for both meta-training configurations. After 20 adaptation steps, the model trained with a single inner loop update reaches an average MSE of 0.0033 mm² (MAE of 0.048 mm), while the model trained with five inner loop updates approaches an average MSE of 0.0027 mm² (MAE of 0.044 mm). Although the overall predictive performance of MAML and Reptile is comparable, Reptile achieves this accuracy with lower computational cost due to its simpler update rule. While this difference is not critical for the relatively small datasets used in this study, it may become more impactful when training on larger or more complex L-DED datasets. In terms of time complexity, which reflects the computation time to run an algorithm, Reptile requires $O(kp)$, where $k$ is the number of inner loop updates and $p$ is the number of model parameters. In contrast, MAML requires up to $O(k^2 p)$ due to the need to differentiate through the inner loop updates, as described in Section 2. These differences become more noticeable as the number of gradient steps, network size, or dataset size increases.

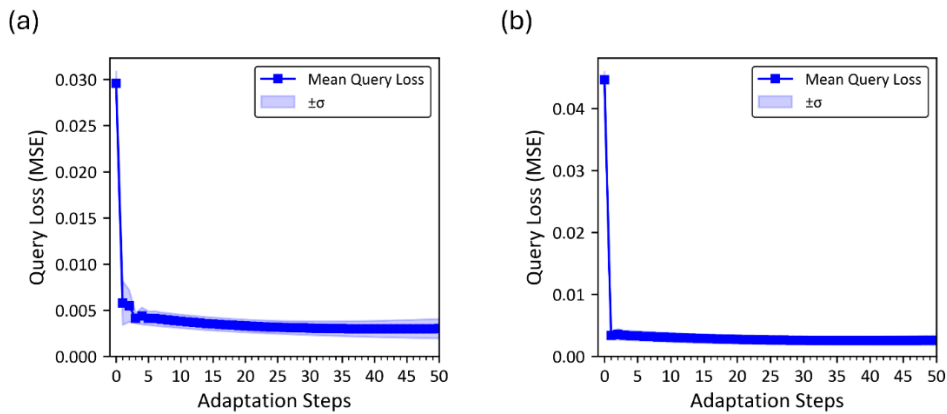

**Fig. 10.** MSE versus adaptation steps for (a) single gradient step and (b) five gradient steps.

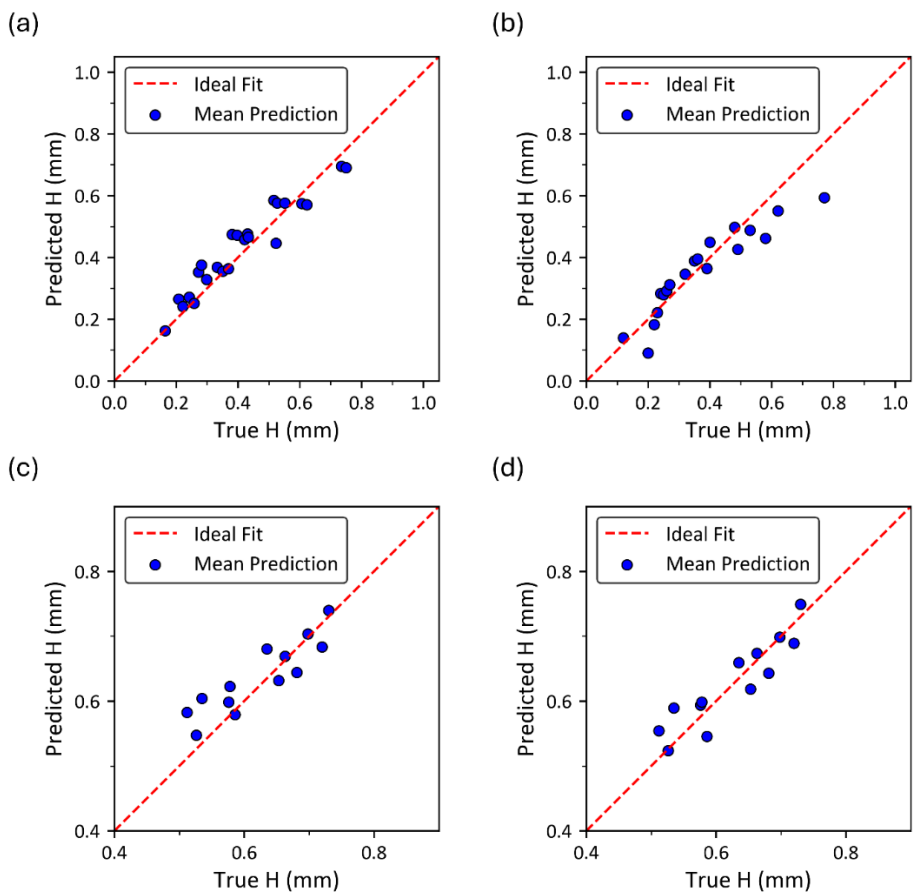

**Fig. 11.** Generalizability of Reptile across different target tasks. Predicted versus true bead height for (a) Task 1 using 20% of the task data as the support set, (b) Task 2 using 20% of the task data as the support set, (c) Task 3 using 20% of the task data as the support set, and (d) Task 3 using 50% of the task data as the support set. (a), (b) and (c) are trained with five inner loop steps and twenty adaptation steps. (d) is trained with five inner loop steps and 400 adaptation steps.

Similar to MAML, the generalizability of Reptile across powder-only L-DED tasks is evaluated by retraining the model with the original target task included in the meta-training set and testing its adaptation on two previously unseen tasks. The same target tasks used in the MAML analysis (Tasks 2 and 3) and the same support set shuffles are used here to enable direct comparison. As shown in Fig. 11, Reptile maintains strong performance on Task 2, achieving an $R^2$ score of 0.83 with five inner loop updates and twenty adaptation steps. When evaluated on Task 3, however, the $R^2$ score decreases to 0.67, reflecting the challenge posed by the smaller dataset and the reduced number of support samples available for adaptation. Increasing the support set fraction from 20% to 50% alone did not substantially improve Reptile's performance on this task, while increasing the number of adaptation steps did improve the performance indicating that the model is still converging. The $R^2$ score continued to improve, reaching 0.80 after 400 adaptation steps. These observations align with the trends observed for MAML, reinforcing the influence of task size and support set composition on model adaptability, and highlighting that additional adaptation steps can partially compensate for limited support data in more challenging target tasks.

### 4.3 Performance of MAML and Reptile for wire-only L-DED

To further evaluate the adaptability of the meta-learning framework across different L-DED process modalities, the performance of the MAML and Reptile algorithms is next examined using wire-only feedstock conditions. While Section 4.2 demonstrated the models' ability to generalize across powder-based tasks, wire-fed L-DED introduces distinct thermal, geometric, and deposition rate characteristics that differ from powder delivery and therefore represent a meaningful test of model robustness. In this section, the models are evaluated using multiple support set shuffles drawn from the wire-only dataset, following the same few-shot adaptation protocol employed for the powder-only tasks. Task 4, previously introduced in table 1, is selected as the baseline target task, while Tasks 5 and 6 will later be used to assess the broader generalizability of the models.

#### 4.3.1 MAML performance, convergence behavior, and generalizability trends

MAML is first evaluated using five inner loop gradient steps and twenty adaptation steps. Using only 20% of the available samples as the support set, the model yields an average $R^2$ of -0.71, with corresponding average MSE and MAE values of 0.305 mm$^2$ and 0.382 mm, respectively, indicating that the baseline initialization learned from the training data does not transfer effectively to wire-fed L-DED tasks. Increasing the support set fraction to 50% improved performance moderately, raising the average $R^2$ to 0.63 (MSE = 0.066 mm$^2$, MAE = 0.199 mm). Evaluating the model on Tasks 5 and 6 produced average $R^2$ values of 0.28 and 0.21 (50% support, 100 adaptation steps), further highlighting the challenge of adapting the meta-model to wire-fed L-DED conditions.

This limited transferability can be attributed to two primary factors. First, the sensitivity of bead height to process parameters is noticeably weaker in wire-fed L-DED compared to powder-fed L-DED, a trend that is already evident from the bead height distributions shown in Fig. 1 (Section 3.1). In powder-fed L-DED systems, bead height responds strongly to changes in energy density because the amount of melted powder varies with laser power, scan speed, and powder mass flow rate. In contrast, in wire-fed L-DED systems, the total delivered mass remains nearly constant, and variations in heat input induce more subtle changes in melt pool geometry, especially bead height. As a result, the mapping between process parameters and bead height is flatter and inherently more difficult to learn from limited support data. Second, the three wire-only target tasks differ significantly in size (22, 44, and 16 samples), which directly impacts the model's adaptability. The performance decrease on the 44 samples task, for example, can be attributed to its removal from the meta-training set, leaving the model without exposure to its dense parameter map. Likewise, the smallest task provides a smaller support set, limiting the model's ability to estimate task-specific gradients during adaptation.

To further isolate the factors constraining model adaptation and to better align the meta-training tasks distribution with the wire-fed target tasks, MAML is retrained using only the wire-only and wire–powder hybrid tasks, with all powder-only tasks removed. In addition, large tasks are split into smaller subtasks to balance the support set size across tasks. Fig. 12 compares bead height predictions under the two training strategies. The model achieved an average $R^2$ of 0.68 (MSE = 0.007 mm$^2$, MAE = 0.059 mm) using 50% support and 20 adaptation steps for Task 4 under the revised meta-training strategy, which represents a comparable performance to the 0.63 $R^2$ obtained under the previous training configuration. Increasing the adaptation steps to 50 improved performances, yielding an average $R^2$ of 0.73 (MSE = 0.006 mm$^2$, MAE = 0.053 mm). These results support the hypothesis that wire-fed and powder-fed L-DED exhibit fundamentally different process–geometry relationships, and that meta-training must include tasks with similar physical behavior to the intended target.

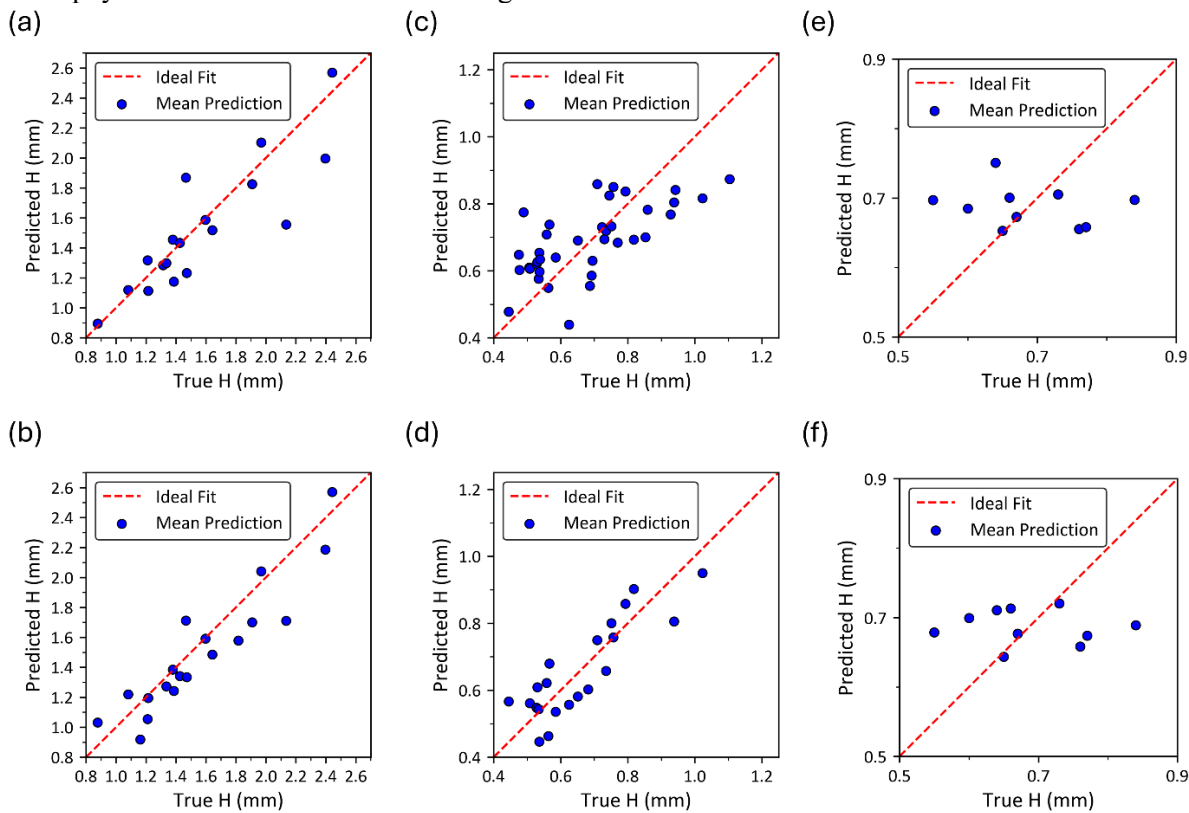

**Fig. 12.** Performance and generalizability of MAML on three different target tasks. (a), (c), and (e) illustrate predictions for Tasks 4, 5, and 6, respectively, trained on the entire dataset. (b), (d), and (f) illustrate predictions for Tasks 4, 5, and 6, respectively, trained on wire-only tasks. All tasks are trained with 5 inner loop steps and 1000 adaptation steps. Here, "H" represents bead height.

When tested on Tasks 5 and 6, the model achieved average $R^2$ values of 0.65 and 0.12 using 50% support and 50 adaptation steps. Increasing the number of adaptation steps improved performance for the first task, with the average $R^2$ rising to 0.7 after 100 adaptation steps. In contrast, performance on the second task did not benefit from additional adaptation. In fact, the average $R^2$ degraded when the number of adaptation steps exceeded 100. This divergence suggests that the second task differs substantially from the tasks used during meta-training. Indeed, this task involves significantly higher laser powers and scan speeds, and uses a wire fed at a 2.5° inclination relative to the x-axis. These conditions are far from the conditions under which the rest of the training tasks are obtained. Consequently, the model's initialization is not well suited for this task, and extended adaptation steps overfit to the small support set rather than improving generalization.

Collectively, these findings highlight the importance of task selection in meta-training. When the process conditions of the target task differ from the conditions represented in the meta-training set, as is the case between wire-fed and powder-fed L-DED, the learned initialization may not provide a suitable starting point for rapid adaptation. As a result, performance becomes highly dependent on the size and representativeness of the support set, the similarity between training and target tasks, and the alignment of process parameter ranges.

### 4.3.2 Reptile performance, convergence behavior, and generalizability trends

Reptile's performance on the wire-only L-DED tasks follows similar trends to those observed for MAML. The model is first evaluated after meta-training on the full dataset, which includes wire-fed, powder-fed, and wire–powder hybrid tasks.

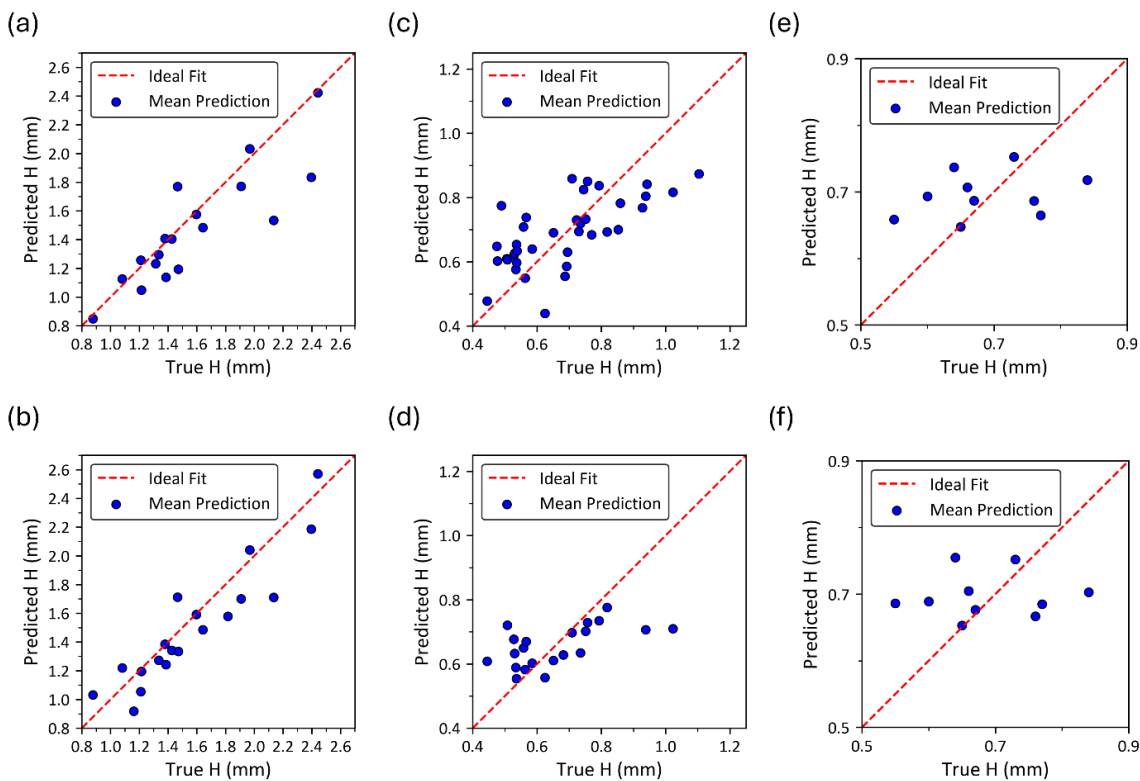

**Fig. 13.** Performance and generalizability of Reptile on three different target tasks. (a), (c), and (e) illustrate predictions for Tasks 4, 5, and 6, respectively, trained on the entire dataset. (b), (d), and (f) illustrate predictions for Tasks 4, 5, and 6, respectively, trained on wire-only tasks. All tasks are trained with 5 inner loop steps and 1000 adaptation steps. Here, "H" represents bead height.

When evaluated on Task 4 using five inner loop updates and twenty adaptation steps, the model produces an average $R^2$ of –0.79. Increasing the support set fraction to 50% improves the average $R^2$ to 0.17. The remaining two target tasks, 5 and 6, yield $R^2$ values of –0.26 and 0.24 under the same conditions, confirming the limited transferability observed previously for MAML in Section 4.3.1. Increasing the adaptation steps to 1000 improved the $R^2$ scores of the first two tasks to 0.66, and 0.43 while the third task's average score decreased to -0.01. To address this issue, Reptile is retrained using the same mitigation strategy implemented for MAML where wire-only and hybrid wire–powder tasks are included in meta-training, and all powder-only tasks are removed. Fig. 13. compares the predictions of bead height under both configurations. Under the revised configuration, the three target tasks evaluated using five inner loop updates, twenty adaptation steps, and a 50% support set achieve R² scores of 0.19, -0.59, and –0.78. Increasing the number of adaptation steps improves performance for the first two tasks, with average R² values rising to 0.77 and 0.15 after 1000 adaptation steps. The third task shows much weaker improvement, reaching only –0.25 even after additional adaptation steps. The corresponding averaged MSE values for the three tasks are 0.050 mm², 0.017 mm², and 0.010 mm², and the MAE values are 0.180 mm, 0.101 mm, and 0.082 mm.

These results reinforce the conclusion that Task 6 lies outside the distribution of the training tasks and that both meta-learning algorithms struggle to adapt to it under few-shot constraints. Reptile achieves better performance than MAML for Task 4 after extended adaptation, while underperforms on Task 5. Yet, both algorithms show similar sensitivity to task similarity, support set size, and alignment of process parameter ranges. In addition, the wire-only results confirm that bead height in wire-fed L-DED responds differently to changes in process parameters compared to powder-fed L-DED, resulting in weaker and more subtle height variations. This difference further emphasizes the need for selecting meta-training tasks that reflect the physical behavior of the target domain to achieve a suitable initialization and faster adaptation during meta-testing.

### 4.4 Performance of MAML and Reptile for wire-powder L-DED

To assess the capability of the meta-learning framework to generalize and adapt to hybrid L-DED conditions, the performance of the MAML and Reptile models is next evaluated on a wire–powder target task (Task 7), which introduces additional complexity due to the simultaneous deposition of wire and powder feedstocks. Evaluating MAML and Reptile in this setting therefore provides a more rigorous test of their ability to extract shared structures across tasks and rapidly adapt to new process conditions with very limited data. As in the previous sections, five independent shuffles of the same target task are used, where 20% of the data is randomly selected as the support set for inner loop adaptation, and the remaining samples serve as the query set for evaluation.

#### 4.4.1 MAML performance, convergence behavior, and generalizability trends

MAML achieves strong performance on the wire–powder target task. Using five inner loop updates, twenty adaptation steps, and a 20% support set (7 samples), the model reaches average R² of 0.81, with corresponding average MSE and average MAE values of 0.010 mm² and 0.074 mm. These values are comparable to the powder-only results and better than the wire-only results, indicating that MAML can generalize to hybrid wire–powder conditions despite the increased process complexity. Varying the number of inner loop and adaptation steps confirms the same trends observed in previous sections, where increasing the number of inner loop gradient steps from one to five and the number of adaptations from five to twenty improves the average R² from 0.57 to 0.81. Furthermore, the MSE loss convergence behavior matches earlier observations, with a rapid reduction in error during the first few adaptation steps and reaching near full convergence within 10–20 steps. Overall, these results show that MAML can accurately predict bead height for a previously unseen wire–powder task using only seven support samples.

The model is then retrained twice, using either wire-fed tasks or powder-fed tasks as the meta-training set. These two additional configurations allow direct comparison of how each task type contributes to the model's initialization and its ability to adapt to the wire–powder target task. Fig. 14 illustrates MAML's performance when trained using the three task combinations, while Table 3 summarizes the model's performance metrics under each configuration. The model trained on the full dataset and the model trained on powder-only tasks achieve similar performance (R² = 0.81, MSE = 0.011 mm², MAE = 0.084 mm), whereas the model trained only on wire-fed tasks performs poorly under the same model hyperparameters using five inner loop updates, twenty adaptation steps, and a 20% support set (R² = -0.21, MSE = 0.068 mm², MAE = 0.208 mm). Increasing the number of adaptation steps to 1000 substantially improves the performance of the model trained with wire-fed L-DED tasks (R² = 0.77, MSE = 0.013 mm², MAE = 0.081 mm), indicating that the model can eventually adapt but requires many more iterations to compensate for the mismatch between training and target task types. In contrast, the models trained with all tasks or powder-only tasks do not benefit from additional adaptation steps, confirming that they reach a suitable initialization much earlier. These results suggest that the bead height trends in the wire–powder task are more similar to those observed in powder-only L-DED, and that including powder-based tasks during meta-training is important for achieving a good initialization and rapid adaptation to the hybrid process.

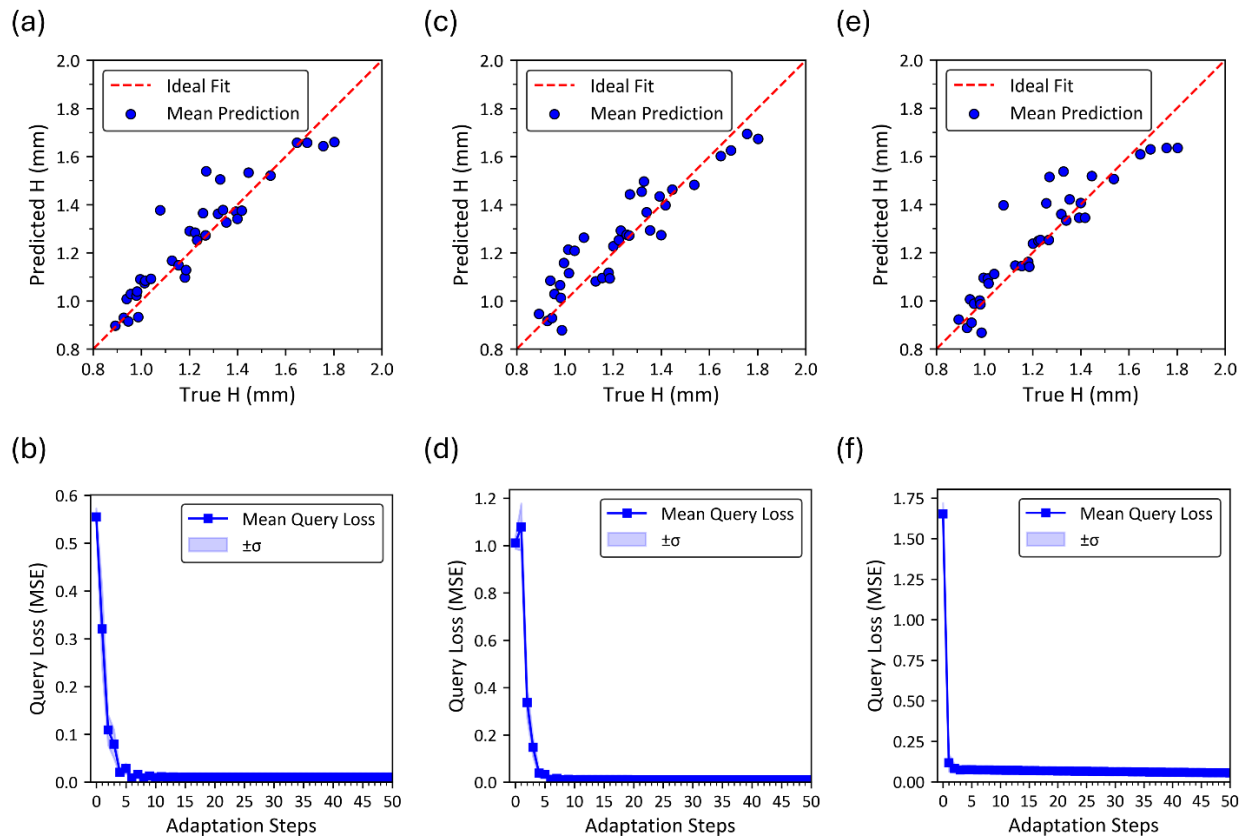

**Fig. 14.** Performance and convergence trends of MAML tested on the wire-powder task. (a), (c), and (e) illustrate bead height predictions of MAML models trained with entire dataset, powder-only tasks, and wire-only tasks. (b), (d), and (f) illustrate convergence of MAML models trained with entire dataset, powder-only tasks, and wire-only tasks. All models are trained with five inner loop steps. (a) and (c) are generated with only 20 adaptation steps while (e) is generated with 1000 adaptation steps. Here, "H" represents bead height.

**Table 3:** Regression performance metrics of MAML model tested on the wire-powder task under different training configurations. All models are trained with support sets that represent 20% of the target task (7 samples).

| Training tasks | Inner loop steps | Adaptation steps | Average Pearson correlation factor | Average $R^2$ score | Average MSE | Average MAE |
|---|---|---|---|---|---|---|
| Wire-only tasks | 1 | 5 | 0.05 | -0.34 | 0.075 | 0.217 |
| | 5 | 20 | 0.23 | -0.21 | 0.068 | 0.208 |
| | 5 | 1000 | 0.90 | 0.77 | 0.013 | 0.081 |
| | 1 | 5 | 0.86 | 0.58 | 0.024 | 0.125 |
| | 5 | 20 | 0.92 | 0.81 | 0.011 | 0.084 |
| Powder-only tasks | 5 | 1000 | 0.92 | 0.81 | 0.011 | 0.072 |
| All tasks | 1 | 5 | 0.82 | 0.57 | 0.024 | 0.125 |
| | 5 | 20 | 0.92 | 0.81 | 0.010 | 0.074 |
| | 5 | 1000 | 0.91 | 0.78 | 0.012 | 0.076 |

### 4.4.2 Reptile performance, convergence behavior, and generalizability trends

Reptile also performs well on the wire–powder target task, though it requires more inner loop steps or more adaptation steps to reach its best performance. With five inner loop updates and twenty adaptation steps, the model achieves average $R^2$ of 0.52 (MSE = 0.027 mm$^2$, MAE = 0.134 mm). Increasing the number of adaptation steps improves performance, reaching an $R^2$ of 0.80 after 150 adaptation steps. Alternatively, the same $R^2$ can be achieved using twenty adaptation steps by increasing the number of inner loop updates from five to thirty. These trends indicate that the meta-training procedure provides a useful initialization, but Reptile requires deeper task-specific adaptation than MAML to reach optimal performance. This behavior is expected because MAML explicitly differentiates through the inner loop updates, allowing each gradient step to carry more information about how the model should adapt, whereas Reptile relies on first order updates that approximate this behavior and therefore benefit more from additional adaptation. They also show that the powder-only, wire-only, and wire–powder datasets contain transferable information, as the model successfully predicts bead height for the hybrid process despite not being exposed to any wire–powder tasks during meta-training, as illustrated in Fig. 15. The regression performance metrics for the three training configurations are presented in Table 4.

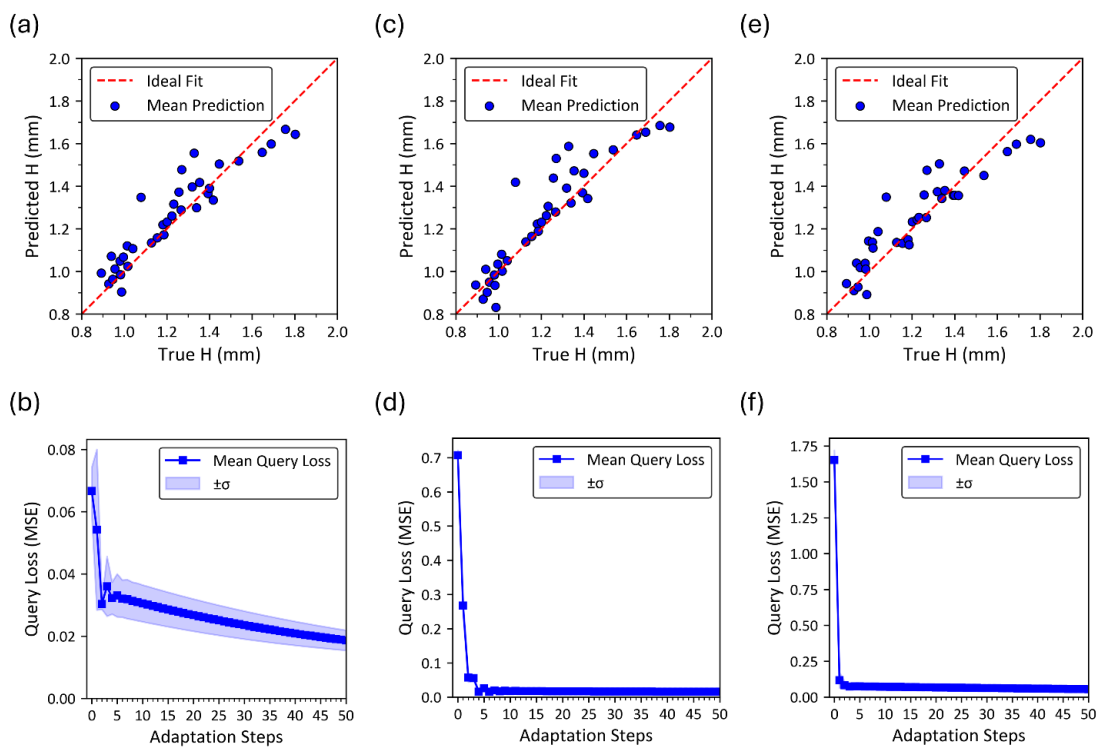

**Fig. 15.** Performance and convergence trends of Reptile tested on the wire-powder task. (a), (c), and (e) illustrate bead height predictions of Reptile models trained with entire dataset, powder-only tasks, and wire-only tasks. (b), (d), and (f) illustrate convergence of Reptile models trained with entire dataset, powder-only tasks, and wire-only tasks. All models are trained with five inner loop steps. (a) and (c) are generated with 150 adaptation steps while (e) is generated with 500 adaptation steps. Here, "H" represents bead height.

Reptile's convergence behavior, shown in Fig. 15(b), (d), and (f), match earlier observations in this study. The error decreases sharply during the first few adaptation steps, after which improvement becomes gradual or plateaus depending on the task. Similar to MAML, retraining Reptile using only powder-only or only wire-only tasks reveals the influence of meta-training composition. The powder-only model performs comparably to the fully trained model, whereas the wire-only model requires many more adaptation steps to reach its best performance. This further supports the conclusion that bead height trends in the wire–powder task is more closely aligned with powder-fed L-DED behavior and that including powder-based tasks during meta-training yields a better initialization for rapid adaptation.

**Table 4:** Regression performance metrics of Reptile model tested on wire-powder task under different training configurations. All models are trained with support sets that represent 20% of the target task (7 samples).

| Training tasks | Inner loop steps | Adaptation steps | Average Pearson's correlation factor | Average $R^2$ score | Average MSE | Average MAE |
|---|---|---|---|---|---|---|
| Wire-only tasks | 5 | 20 | 0.23 | -0.21 | 0.068 | 0.208 |
|  | 5 | 150 | 0.88 | 0.48 | 0.029 | 0.139 |
|  | 5 | 500 | 0.90 | 0.74 | 0.015 | 0.090 |
| Powder-only tasks | 5 | 20 | 0.89 | 0.69 | 0.017 | 0.093 |
|  | 5 | 150 | 0.91 | 0.75 | 0.014 | 0.084 |
|  | 5 | 500 | 0.91 | 0.77 | 0.013 | 0.080 |
| All tasks | 5 | 20 | 0.78 | 0.52 | 0.027 | 0.134 |
|  | 5 | 150 | 0.92 | 0.80 | 0.011 | 0.079 |
|  | 5 | 500 | 0.92 | 0.80 | 0.011 | 0.075 |

### 4.5 Implications for geometry prediction in L-DED

The results obtained across powder-based, wire-based, and wire-powder L-DED tasks demonstrate that meta-learning offers several important advantages for bead geometry prediction compared with conventional machine learning models. In L-DED, data scarcity, heterogeneity across machines and process conditions, and variability in task distributions pose persistent challenges to generalizable predictive modeling. The findings of this study show that both MAML and Reptile explicitly address these challenges by learning initialization parameters that can rapidly adapt to new process conditions using only a few support samples. This characteristic is particularly valuable in L-DED workflows, where experiments are costly, parameter spaces are broad, and accessible datasets are typically small and fragmented.

A key implication is that meta-learning provides a practical mechanism for few-shot prediction of bead geometry under new operating regimes. Whereas a base model trained directly on a single dataset tends to overfit to its local characteristics and struggles when applied to an unseen task, meta-trained models learn shared structural relationships across tasks. When confronted with a new L-DED condition, such as a different material, power level, feedstock delivery mode, or machine configuration, the meta-learned model begins from a parameter state already aligned with the empirical trends common to L-DED processes. As shown in Sections 4.2 and 4.3, this enables accurate adaptation with very limited data, such as using only three to five samples. Conventional models lack this adaptability as retraining them from scratch requires many more samples.

To establish a baseline for comparison with the meta-learning models, a conventional feedforward neural network is trained using the same architecture and hyperparameters employed throughout this study. The model consists of three fully connected layers with 64 neurons per layer, trained for 100 epochs using the Adam optimizer with a learning rate of 0.001. The network is evaluated under two training configurations to assess both its capacity for generalization across tasks and its performance when focused on a specific task. In the first configuration, the network is trained with all L-DED tasks together with 20% of the samples from the target task. In the second configuration, the network is trained solely on 20% of the target task itself, mimicking a few-shot learning setting but without any meta-learning mechanism. Fig. 16 presents the average predicted bead height versus the true bead height across five independent shuffles of the target task and training configuration.

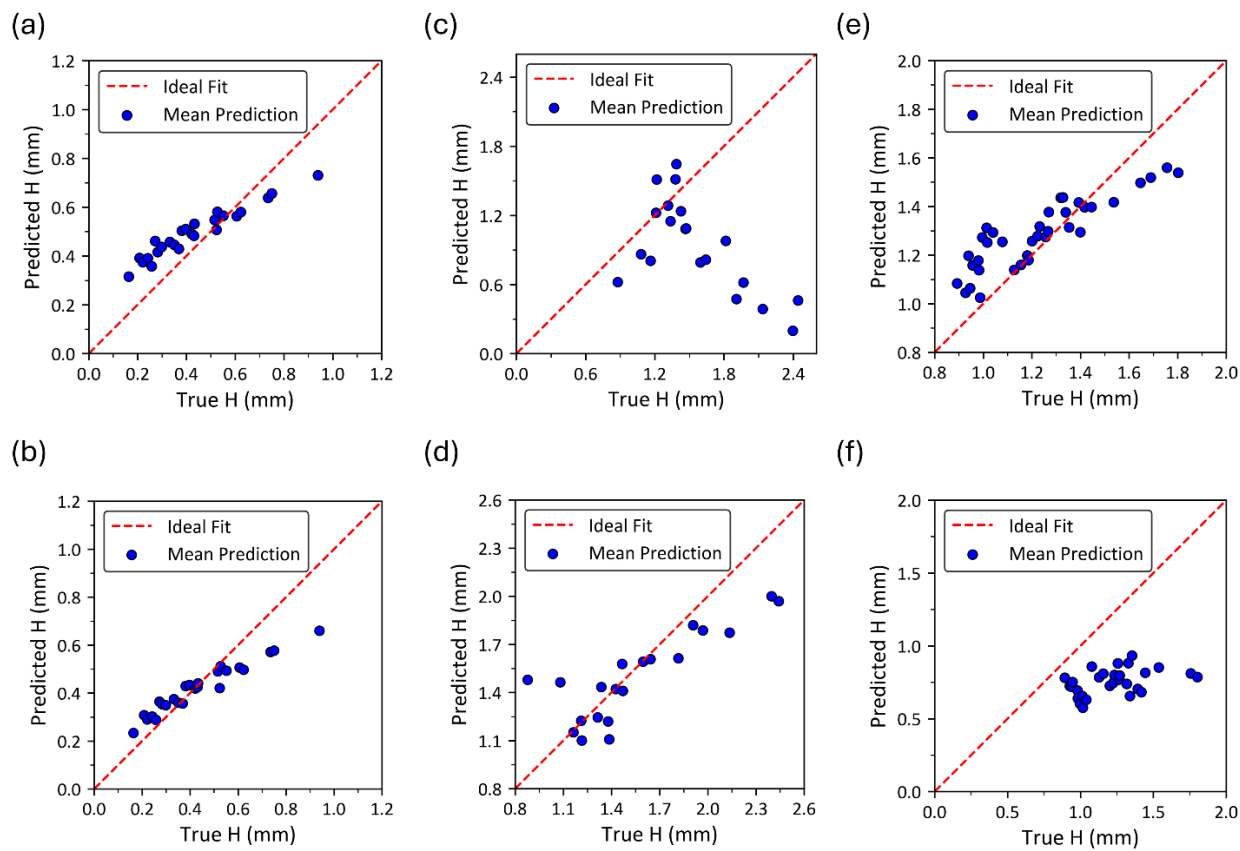

**Fig. 16.** Average predicted bead height versus true bead height across five independent shuffles for conventional feedforward neural networks evaluated on Tasks 1, 4, and 7. (a), (c), and (e) represent models trained on all L-DED tasks together with 20% of the target task data, while (b), (d), and (f) represent models trained using only 20% of the target task data. Here, "H" represents bead height.

The quantitative metrics reported in Table 5 further highlight the limitations of the feedforward neural network relative to the meta-learning approaches. When trained on the entire dataset which includes aggregated tasks representing various printing conditions, the network performs poorly across all tasks, exhibiting strongly negative $R^2$ values and large prediction errors. This outcome indicates that the network does not extract transferable structure from heterogeneous L-DED data and instead becomes biased toward the dominant trends of the training set, leading to severe mismatches when applied to a new task. Training the model only on 20% of the target task improves performance substantially. However, even in this best-case scenario, the network remains markedly less accurate than both MAML and Reptile. These results reinforce the core findings of this study: while standard neural networks must essentially relearn each new task from scratch, meta-learning produces an initialization that generalizes across tasks and adapts effectively with only a few support samples. Consequently, meta-learning provides a more robust and data-efficient framework for geometry prediction in L-DED processes, particularly in scenarios where task variability is high and data availability is limited.

**Table 5:** Regression performance metrics of the conventional feedforward neural network under different training configurations, including training on all L-DED tasks together with 20% of the target task and training exclusively on the 20% of the target task.

| Target task | Training configuration | Average Pearson correlation factor | Average $R^2$ score | Average MSE | Average MAE |
|---|---|---|---|---|---|
| Task 1 (powder-only) | All L-DED tasks + 20% of the target task | 0.97 | 0.65 | 0.013 | 0.10 |
| Task 3 (powder-only) | All L-DED tasks + 20% of the target task | 0.86 | -17.42 | 0.10 | 0.31 |
| Task 4 (wire-only) | All L-DED tasks + 20% of the target task | -0.47 | -3.4 | 0.84 | 0.65 |
| Task 5 (wire-only) | All L-DED tasks + 20% of the target task | 0.13 | -0.88 | 0.56 | 0.17 |
| Task 7 (wire-powder) | All L-DED tasks + 20% of the target task | 0.45 | -3.9 | 0.28 | 0.48 |
| Task 1 (powder-only) | 20% of the target task | 0.75 | 0.36 | 0.02 | 0.75 |
| Task 3 (powder-only) | 20% of the target task | -0.19 | -0.80 | 0.01 | 0.81 |
| Task 4 (wire-only) | 20% of the target task | 0.58 | 0.26 | 0.14 | 0.30 |
| Task 5 (wire-only) | 20% of the target task | 0.24 | -1.01 | 0.06 | 0.18 |
| Task 7 (wire-powder) | 20% of the target task | 0.78 | 0.33 | 0.04 | 0.15 |

## 5. Conclusions

This paper presents a comprehensive investigation into few-shot bead height prediction for L-DED tasks using a meta-learning-based framework. Experimental datasets from multiple sources, including peer-reviewed literature and in-house experiments, are compiled and organized as learning tasks. These learning tasks reflect variations in materials, feedstock delivery modes, and processing conditions in L-DED. Gradient-based meta-learning algorithms, i.e., MAML and Reptile, are implemented using a feedforward neural network and systematically evaluated across powder-only, wire-only, and hybrid wire–powder L-DED tasks. Prior to application on L-DED data, the correctness and stability of the meta-learning framework are rigorously validated by reproducing established literature benchmarks, including the sinusoidal regression task reported in [58] and a previously reported meta-learning framework for bead geometry prediction [60]. This validation step ensures that the inner and outer loop optimization structures, task generation procedures, and adaptation mechanisms are correctly implemented. The validated framework is then applied to the compiled L-DED datasets, where MAML and Reptile are evaluated using multiple experimental configurations. The main findings of this study can be summarized as follows:

1) Both MAML and Reptile demonstrated strong few shot learning capability, achieving accurate bead height predictions on unseen target tasks using as little as 20% of the target task data and, which in practice corresponded to as few as three to seven support samples depending on task size. For powder-only and wire–powder L-DED tasks, the meta-learned models consistently achieved high predictive accuracy ($R^2$ values up to ~0.9), substantially outperforming conventional feedforward neural networks trained either on the entire compiled datasets or directly on 20% of the target task data. This confirms that meta-learning effectively captures transferable process–geometry relationships that conventional supervised learning might fail to generalize.

2) The comparative analysis between MAML and Reptile revealed that while both algorithms achieve comparable peak performance, they exhibit distinct trade-offs. MAML generally converges faster during adaptation and requires fewer gradient steps to reach optimal accuracy, owing to its use of second order gradients to optimize the task adaptation process. Reptile, in contrast, relies on a first order gradient-based update that approximates this behavior without computing second order derivatives, making it more computationally efficient and easier to scale to larger models or datasets. These observations highlight the key considerations that must be considered when selecting a meta-learning algorithm, including the size of the dataset, available computational resources, and the optimal number of adaptation steps in practical L-DED applications.

3) The study shows that task similarity and the composition of the meta-training dataset play a critical role in successful knowledge transfer. Meta-models trained on powder-based tasks generalized well to wire–powder targets, whereas models trained exclusively on wire-fed tasks struggled to adapt to the same tasks. Wire-only L-DED tasks, in particular, posed a greater challenge due to weaker sensitivity of bead height to process parameters. These results highlight the importance of aligning the meta-training task distribution with the dominant process–geometry trends of the target domain to enable efficient and reliable adaptation.

4) Baseline results obtained with a conventional feedforward neural network demonstrate the limitations of standard supervised learning in heterogeneous and data-scarce L-DED settings. Using the same model architecture, conventional models exhibited poor generalization and unstable performance, reinforcing the advantage of meta-learning as a robust and data-efficient alternative for bead geometry prediction.

This work lays out the foundation for several promising directions for future research. Extending the proposed framework to simultaneously predict multiple geometric features, such as bead width, cross-sectional area, and penetration depth, would enable a better description of deposition quality, which is essential for reliable process optimization and control in L-DED [40], [60]. Incorporating additional input modalities, such as in-situ thermal images, melt pool monitoring signals, or temporal process histories, could further enhance predictive accuracy and robustness under complex operating conditions [74]. Finally, validating and integrating the proposed meta-learning framework in experimental settings, either in open-loop or closed-loop process control settings, would represent an important step toward practical deployment in industrial L-DED systems [75].

**CRediT authorship contribution statement**
Abdul Malik Al Mardhouf Al Saadi: Data curation, Formal analysis, Investigation, Methodology, Software, Validation, Visualization, Writing – original draft, Writing – review & editing. Amrita Basak: Writing – review & editing, Supervision, Resources, Project administration, Funding acquisition, Conceptualization.

**Funding information**
This work is supported by the Department of Energy through grant number DE-NE0009253. Any opinions, findings, and conclusions in this paper are those of the authors and do not necessarily reflect the views of the supporting institution.

**Declaration of Generative AI and AI-assisted technologies in the writing process**
During this work's preparation, OpenAI ChatGPT has been used to improve grammar and readability.

**Declaration of competing interest**
The authors declare that they have no known competing financial interests or personal relationships that could have appeared to influence the work reported in this paper.

**Acknowledgement**
The authors would like to thank Kun-Hao Huang for having valuable discussions on data science and machine learning.

**Data availability**
Data will be made available upon reasonable request.